\documentclass[journal]{IEEEtran}

\usepackage{times}
\usepackage{epsfig}
\usepackage{graphicx}
\usepackage{amsmath}
\usepackage{amssymb}
\usepackage{makecell,rotating}
\usepackage{booktabs}
\usepackage{multirow}
\usepackage[pagebackref=true,breaklinks=true,letterpaper=true,colorlinks,bookmarks=false]{hyperref}
\usepackage{algorithm}  
\usepackage{algpseudocode}  
\begin{document}

\title{Birds of A Feather Flock Together: Category-Divergence Guidance for Domain Adaptive Segmentation}

\author{Bo~Yuan,
		Danpei Zhao, %~\IEEEmembership{Member,~IEEE,}
        Shuai~Shao,
        Zehuan~Yuan,
        and~Changhu~Wang%~\IEEEmembership{Life~Fellow,~IEEE}% <-this % stops a space

\thanks{Bo Yuan, Danpei Zhao are with the Image Processing Center, School of Astronautics, Beihang University, Beijing 100191, China (e-mail: yuanbobuaa@buaa.edu.cn, zhaodanpei@buaa.edu.cn).}% <-this % stops a space
\thanks{Shuai Shao, Zehuan Yuan, Changhu Wang are with ByteDance AI-Lab, Beijing 100086, China (e-mail: shaoshuai@acm.org, yuanzehuan@bytedance.com, wangchanghu@bytedance.com).}% <-this % stops a space
%\thanks{Manuscript received X X, X; revised X X, X.}
}

% The paper headers
\markboth{Journal of \LaTeX\ Class Files,~Vol.~x, No.~x, August~x}%
{Shell \MakeLowercase{\textit{et al.}}: Bare Demo of IEEEtran.cls for IEEE Journals}

% make the title area
\maketitle

\begin{abstract}
Unsupervised domain adaptation (UDA) aims to enhance the generalization capability of a certain model from a source domain to a target domain. Present UDA models focus on alleviating the domain shift by minimizing the feature discrepancy between the source domain and the target domain but usually ignore the class confusion problem.  In this work, we propose an Inter-class Separation and Intra-class Aggregation (ISIA) mechanism. It encourages the cross-domain representative consistency between the same categories and differentiation among diverse categories.  In this way, the features belonging to the same categories are aligned together and the confusable categories are separated. By measuring the align complexity of each category, we design an Adaptive-weighted Instance Matching (AIM) strategy to further optimize the instance-level adaptation. Based on our proposed methods, we also raise a hierarchical unsupervised domain adaptation framework for cross-domain semantic segmentation task. Through performing the image-level, feature-level, category-level and instance-level alignment, our method achieves a stronger generalization performance of the model from the source domain to the target domain. In two typical cross-domain semantic segmentation tasks, i.e., GTA5$\rightarrow$Cityscapes and SYNTHIA$\rightarrow$Cityscapes, our method achieves the state-of-the-art segmentation accuracy. We also build two cross-domain semantic segmentation datasets based on the publicly available data, i.e., remote sensing building segmentation and road segmentation, for domain adaptive segmentation. \footnote{Part of this work was done while Bo Yuan was an intern at ByteDance AI-Lab. Our code, models and datasets will be available at our formal published version.} \footnote{ $\copyright$2022 IEEE. Personal use of this material is permitted. Permission from IEEE must be obtained for all other uses, in any current or future media, including reprinting/republishing this material for advertising or promotional purposes, creating new collective works, for resale or redistribution to servers or lists, or reuse of any copyrighted component of this work in other works. }
\end{abstract}


\IEEEpeerreviewmaketitle

\section{Introduction}

\IEEEPARstart{S}{emantic} segmentation aims to assign a label to every pixel in the image, which normally requires large-scale pixel-level annotated data for training an applicable model. However, it is extremely time-consuming and labor-intensive to collect data with pixel-level annotations. For example, Cityscapes~\cite{Cityscapes} is a widely-used benchmark dataset and it takes 1.5 hours on average to annotate an image; which sums up to about 7500 hours totally to annotate all 5000 images. However, in comparison, training an applicable semantic segmentation model on the collected data usually takes only several hours.
\begin{figure}
	\centering
	\includegraphics[scale=0.45]{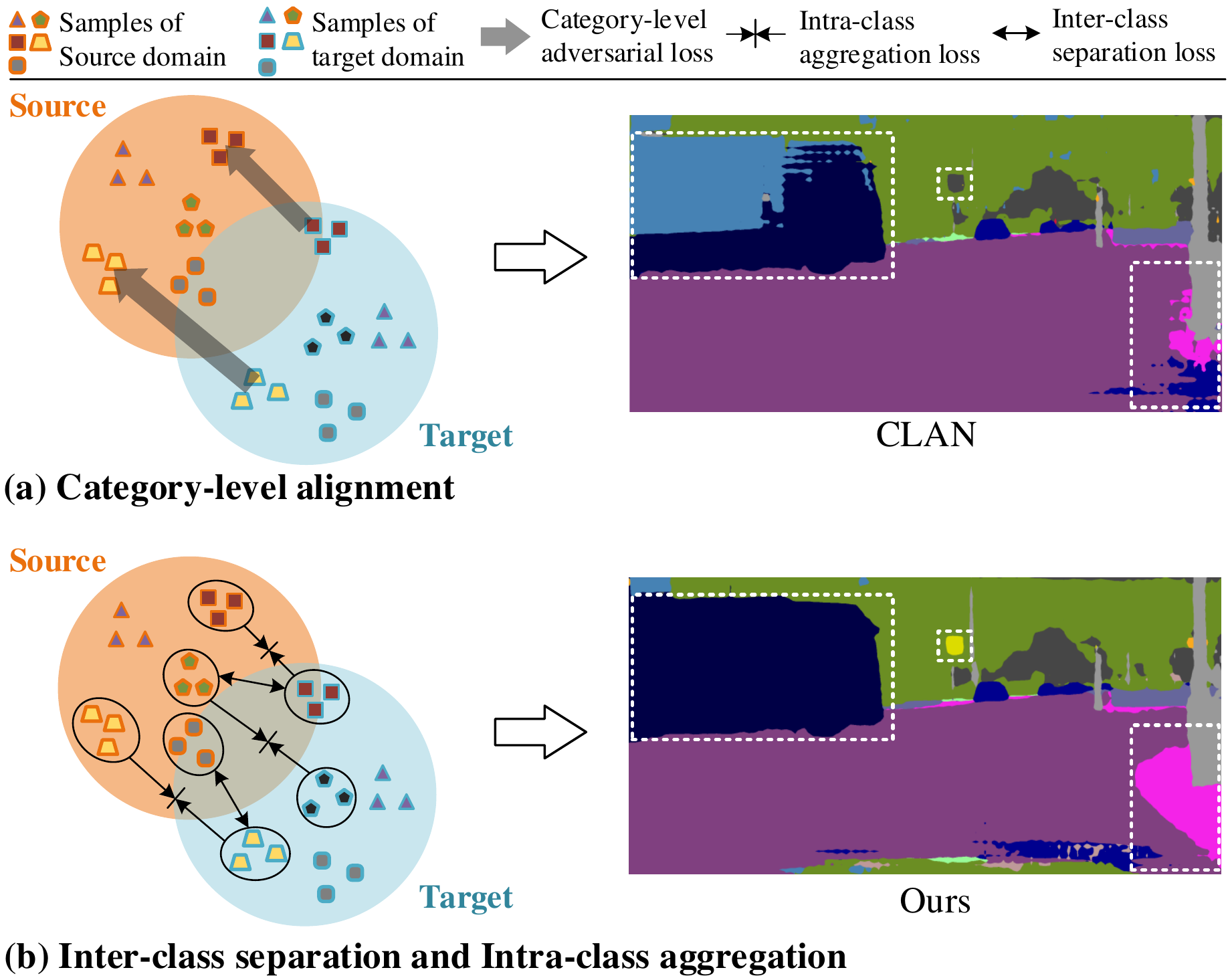}
	\caption{Illustration of the proposed category-divergence guidance for domain adaptive segmentation. (a): Category-level alignment strategy proposed by~\cite{CLAN}. It encourages a category-level joint distribution alignment but is confronted with the class confusion problem. (b): Our proposed inter-class separation and intra-class aggregation mechanism. Our method simultaneously performs feature alignment between the same categories and differentiating among different categories. As shown, the proposed strategy effectively reduces pixels misclassification in cross-domain segmentation task.}
	\label{fig-motivation}
\end{figure}
In recent years, photorealistic data rendered from video games and simulators with pixel-level semantic annotations have been used to train segmentation networks. Normally, the models trained on the synthetic data do not generalize well to realistic target domain. The reason lies in the different data distributions of the different domains, which is typically known as domain shift~\cite{DomainShift}.
Recently, unsupervised domain adaptation (UDA) methods are proposed to address this issue. In such works, a model trained on a source domain dataset with pixel-level segmentation annotations is adapted for an unlabeled target domain. By quantifying the data distribution, domain adaptation approaches~\cite{ADVENT,  CLAN, CyCADA, DeepAA, SimRealJR, BidirectionalLF, DCAN, PyCDA, CBST} are proposed to minimize the feature distribution discrepancy between the source and target domains. A popular domain adaptation choice is to align the image style and feature representations of different domains~\cite{FCAN, AdaptSegNet}. A majority of recent methods~\cite{UDADT, PIT, FADA} explore semantic-level adaptation such as category-level and instance-level alignment. Among this cohort of UDA methods, a common and pivotal approach is minimizing some distance metrics between the source and target feature distributions~\cite{LearningTF, DeepCC, SSFDAN}. Another effective approach, which employs GAN~\cite{GAN} architectures, is to minimize the accuracy of domain prediction. A GAN architecture is usually composed of a generator and a discriminator. The generator extracts features from the input images and the discriminator distinguishes which domain the features are generated from. Through a minimax game between two adversarial networks,  the discriminator can thereby guide the generator to produce the target domain features with a distribution closer to that of the source domain. In recent years, the GAN-based UDA for semantic segmentation has been applied to urban scenes~\cite{AdaptSegNet, FCAN,  Biasetton2019UnsupervisedDA}, aerial remote sensing images~\cite{Benjdira2019UnsupervisedDA, Benjdira2020DataEfficientDA, JPRNet}, LiDAR point cloud~\cite{Jiang2020LiDARNetAB, Yi2020CompleteL}, etc.

Although current adversarial learning methods have led to impressive results~\cite{FCNsITW, LearningTD, CoupledGA, Mauro2020SceneAdaptSD}, there are still limitations can not be ignored: 1) the global adversarial learning approach aligns the global feature distribution in the source and target domains by training a GAN. However, when the generator can perfectly fool the discriminator, the alignment between the source and target domains is still weak for achieving a sufficient segmentation accuracy in target domain because of the low generalization on multiple categories. 2) although category-level domain adaptation approach~\cite{CLAN} and instance-level alignment method~\cite{UDADT} have been proposed to enhance the semantic-level alignment, there is still a problem of pixel aliasing.  Specifically, the features of different categories require to be separated but the alignment strategy lacks such structural information.  For example, the classes such as \emph{sky} and \emph{road} normally vary rarely in color, shape and position in the image, which are easily to be distinguished. While in many situations, the pixels those close to region boundaries of different categories are likely to be misclassified, as shown in Fig.~\ref{fig-motivation}.

To address the limitation of the traditional category-level alignment, we propose an inter-class separation and intra-class aggregation alignment strategy.  By constructing a similarity measure function based on cosine distance, we conduct features alignment between the same categories and features separation among different categories across domains in the meantime.  Through the measuring of the alignment complexity for each category, we design an adaptive weight to further guide the instance-level alignment. 
Our main contributions are summarized as follows:

\begin{itemize}
	\item[$\bullet$]  We propose a category-divergence guidance approach for cross-domain semantic segmentation.  Our model efficiently reduces pixel misclassification by pulling closer feature representations of the same categories and pushing away those belonging to different categories. 
	\item[$\bullet$] We construct a universal UDA framework from multi-level alignments including image level, feature level, category level and instance level, synergistically reducing domain gap.
	\item[$\bullet$] We extend the proposed UDA method to remote-sensing scenes by reforming four representative remote-sensing datasets for cross-domain building segmentation and road segmentation. 
	\item[$\bullet$] The proposed UDA method achieves the state-of-the-art semantic segmentation accuracy on benchmark datasets including street scenes and remote-sensing images.
\end{itemize}

% network
\begin{figure*}[t]
	\centering
	\includegraphics[scale=0.46]{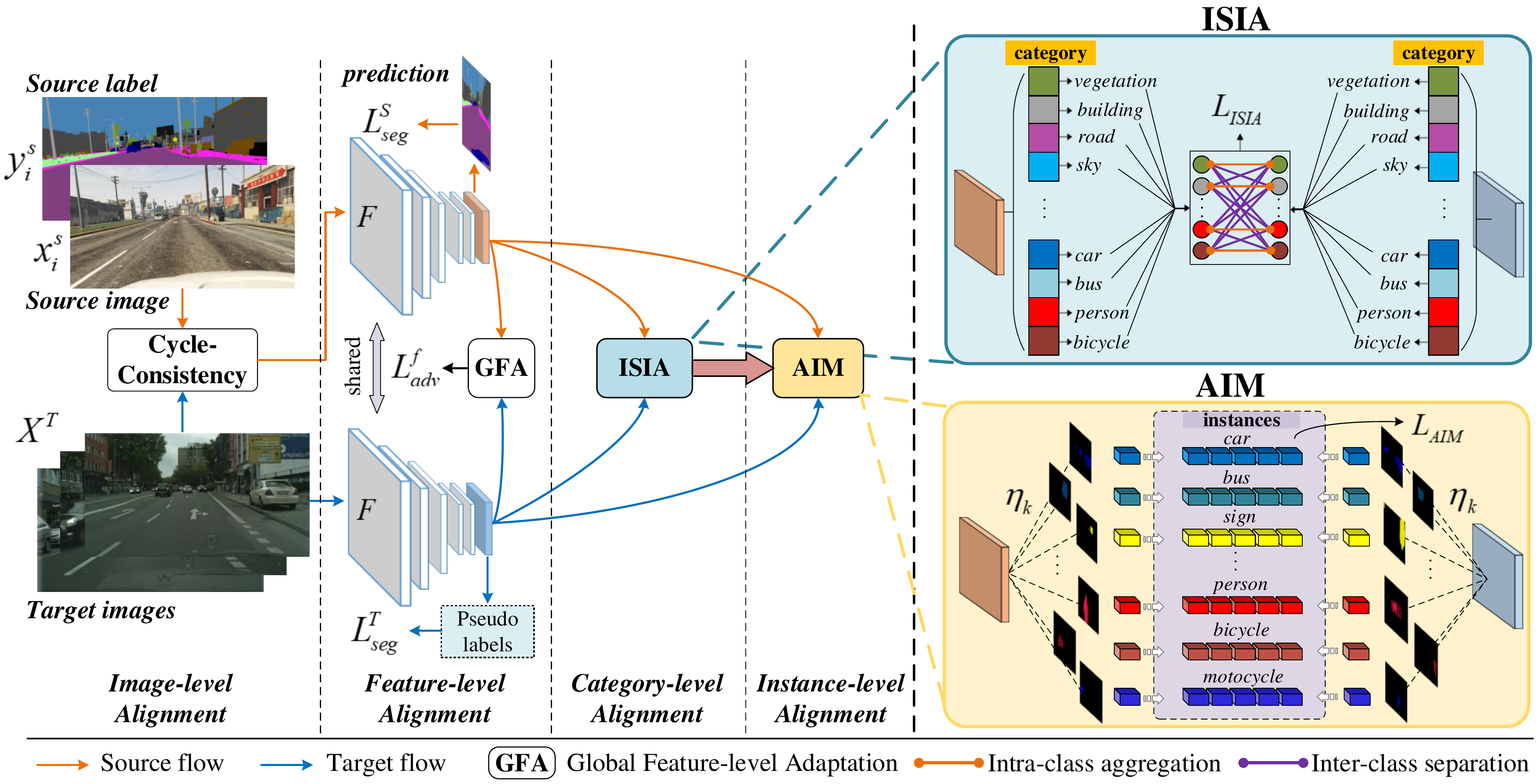}
	\caption{Network Architecture. It consists of image-level, feature-level, category-level and instance-level alignments. GFA: global feature-level adaptation supervised by an adversarial loss. ISIA: inter-class separation and intra-class aggregation module. The features from the source and target domains are split into $N$ classes, $N$ is the number of semantic categories. AIM: adaptive-weighted instance matching. The foreground instances from the source and target domains are aligned by minimizing the cross-domain instance matching loss.}
	\label{fig-network}
\end{figure*}

\section{Related Works}
\textbf{Semantic segmentation.} Semantic segmentation has been significantly boosted with the development of convolutional neural networks. Since~\cite{FCN}, the models based on fully convolutional network (FCN)~\cite{FCN} have grabbed massive attention. Since modeling long-range dependency information is critical for semantic segmentation, extensive efforts have been focused on increasing the receptive field through either using dilated/atrous convolutions~\cite{DeepLabv2, DeepLabv3} or inserting attention modules~\cite{CCNet, Chen2016AttentionTS, OCNet, DANet, PSANet}. Another popular path, \cite{Unet, DeepLabv3+, RefineNet, DFN} adopt encoder-decoder structures that fuse the information in low-level and high-level layers to predict segmentation mask. \cite{PSPNet} utilizes pyramid pooling to aggregate contextual information. \cite{HRNet} starts from a high-resolution subnetwork and gradually adds high-to-low resolution subnetworks one by one to maintain high-resolution representations in the image. Ren et al~\cite{10.1145/3447582} explore neural architecture search (NAS) in semantic segmentation architecture design. Recently,  \cite{SETR, SwinTH} replace traditional convolutional backbones with vision transformers. These methods consider semantic segmentation as a sequence-to-sequence prediction task to dispose the limited receptive fields. However, the advanced performance of these semantic segmentation methods often build on the large amounts of densely annotated images, which are usually difficult to collect.

\textbf{Adversarial learning.} Generative adversarial networks (GANs)~\cite{GAN, Denton2015DeepGI, Oord2016ConditionalIG} learn two networks, i.e., a generator and a discriminator, in a staged zero-sum game fusion to generate images from inputs. The key component enabling GANs is the adversarial constraint, which makes the generated images to be indistinguishable from real images. The GAN-based methods have been widely used in image-level domain mapping. This task focuses on transferring the image style from source domain to target domain, which is popular in image-to-image translation~\cite{FCAN, Salimans2016ImprovedTF,  Chen2018StereoscopicNS, Royer2020XGANUI, Sheng2018AvatarNetMZ} and domain adaptation~\cite{ Gong2018CausalGD, CyCADA, DCAA, PIT}. 

\textbf{Domain adaptation for semantic segmentation.} Many UDA works are designed for classification, like ADDA~\cite{ADDA}, MMD~\cite{MMD}, et al. With the synthetic datasets including GTA5~\cite{GTA5}, SYNTHIA~\cite{SYNTHIA}, Synscapes~\cite{Synscapes} are proposed, UDA for semantic segmentation is also comes to insight. From the adaptation manner, the UDA approaches for semantic segmentation can be divided into image-level, feature-level and label-level methods. 

The image-level adaptation refers to changing the appearance of images such that images from the source domain and the target domain are more visually similar. These methods~\cite{BidirectionalLF, DCAN, FCAN} usually transfer the color, texture, illumination and other stylization factors of images from one domain to another.
%In~\cite{FCAN}, Zhang et al. propose an appearance adaptation network which transfers image appearance while preserving image semantic structures. 
%Li et al.~\cite{BidirectionalLF} utilize CycleGAN~\cite{CycleGAN} with a perceptual loss to preserve the locality of semantic information to conduct the unpaired image-to-image translation. 
Choi et al.~\cite{Choi2019SelfEnsemblingWG} propose a GAN-based self-ensembling data augmentation method for domain alignment. Recently Kang et al.~\cite{PLCA}  propose to build the pixel-level cycle association between source and target pixel pairs and contrastively strengthen their connections to diminish the domain gap. The feature-level  transferring refers to matching the extracted feature distributions between the source and target domain. Deep convolutional neural networks  (CNNs)~\cite{ResNet, VGG, FCN, Huang2017DenselyCC} can extract the features from the source domain and the ones from target domain. However, due to the domain shift~\cite{CLAN}, minimizing the feature distribution discrepancy with GAN~\cite{GAN} structure is a common practice. Tsai et al.~\cite{AdaptSegNet} propose a joint consideration of pixel and feature level adaptation. Li et al.~\cite{CCM} actively select positive source information for training to avoid negative transfer by constructing a content-consistent matching mechanism.
%Hoffman et al.~\cite{FCNsITW} append category statistic constraints to the adversarial model, aiming to improve semantic consistency in target domain.
Wu et al.~\cite{DCAN} raise a channel-wise feature alignment network to close the gap of the channel-wise mean and standard deviation in CNN feature maps. Lv et al.~\cite{PIT} propose a domain-invariant interactive relation transfer strategy to align both the image-level and pixel-level information. The label-level adaptation refers to producing pseudo-labels of the target domain by utilizing the knowledge learned from the source domain, where a self-supervised learning approach~\cite{IntraDA, FADA, UDADT, CAG-UDA, Luo2021GetAF} is usually used. Cai et al.~\cite{DCAA} study adversarial ambivalence by revising the pseudo-labels and emerge the hard adaptation regions.  Besides the single-source setting, multi-source domain adaptation~\cite{Zhao2019MultisourceDA, MultiSourceDA}  for semantic segmentation are also studied. Tasar et al.~\cite{StandardGAN, Tasar2021DAugNetUM} explore domain adaptation in satellite images. %Wang et al.~\cite{Wang2020ClassesMA} design a fine-grained adversarial strategy on class-level alignment.

\section{Preliminaries} 
\textbf{Problem Setting.} Given a source domain dataset with images and pixel-level annotations $\{x_i^s, y_i^s|x_i^s \in X^S, y_i^s \in Y^S\}$, and a target domain with only images $\{x_i^t | x_i^t \in X^T\}$, the goal is to train a model that can produce the pixel-level predictions $\{\hat{y}_i^t\}$ of the target domain images. 

\textbf{Segmentation and adversarial adaptation.}  We focus on training a semantic segmentation model by minimizing the discrepancy between the source and target domains. Firstly, training a model $G$ that distills knowledge from labeled-data in order to minimize the segmentation loss in the source domain:
\begin{equation}
	%\mathcal{L}_{seg}(G) = E(l(G(X^S), Y^S)) 
	\mathcal{L}_{seg}(G) = -\sum_{i=1}^{H\times W}\sum_{k=1}^{N}y_{ik}log p_{ik}
	\label{Eqn1}
\end{equation} 
where $y_{ik}$ and $p_{ik}$ represent the ground truth probability and the predicted probability of class $k$ on pixel $i$, respectively. 
Second,  an adversaries-based UDA method trains $G$ to learn domain-invariant features by fooling a domain discriminator $D$ which is able to distinguish samples belonging to the source or target domains. This goal is achieved by minimaxing an adversarial loss defined in Eqn~(\ref{Eqn2}).
\begin{equation}
	\begin{split}
		\mathcal{L}_{adv}^{f}(G, D) =& -E(log(D(G(X^S))))\\
		&-E(log(1-D(G(X^T))))
	\end{split}
	\label{Eqn2}
\end{equation}
where $E(\cdot)$ represents statistical expectation.

%----------Method Section-----------
\section{Method}
Our model consists of a multi-level alignment framework. Specifically, we conduct the global feature-level alignment together with the proposed category-level and instance-level alignment strategies. The overall network architecture is illustrated in Fig.~\ref{fig-network}. 

\subsection{Global Feature level Adaptation}
\label{sec-4.1}
Firstly, we use cycle-consistency~\cite{CycleGAN, BidirectionalLF} for the unpaired image-to-image translation. This image style transferring process aims to transfer image appearance from the target domain to the source domain, which can be viewed as low-level feature alignment.
To realize the global feature alignment in the output space, the images from the source and target domains are imported to a parameter-shared feature extractor. And we use the spatial layout of the source- and target-domain samples as the input of the discriminator. Following~\cite{UDADT}, we impose a traditional GAN structure on the output space~\cite{AdaptSegNet} to globally minimize the feature distribution discrepancy between the source domain and the target domain. 
A discriminator $D$ will discriminate the generated output by $G$. Here, the generator $G$ is composed of a feature extractor $F$ and a classification head $C$ and $G=F\circ C$. We minimize the feature distribution discrepancy between the source domain and the target domain by optimizing the adversarial target function as follows:
\begin{equation}
	\mathop{min}\limits_{G} \mathcal{L}_{adv}^f(G, D)=-\sum_{x_i^t\in X^T}log(1-D(S(G(x_i^t))))
	\label{Eqn-CE}
\end{equation}
where $S$ is the softmax operation. While the discriminator tries to distinguish which domain the feature is formed by optimizing the discriminator target function as follows:
\begin{equation}
	\begin{split}
		\mathop{min}\limits_{D} \mathcal{L}_{D}(G, D)&=-\sum_{x_i^t\in X^T}log(D(S(G(x_i^t)))) \\
		&-\sum_{x_j^s\in X^S}log(1-D(S(G(x_j^s))))
	\end{split}
	\label{Eqn4}
\end{equation}

\subsection{Divergence-driven Category level Alignment}
\label{sec-ICSA}
The distribution difference of homogeneous features and the confusion of heterogeneous features constitute the key part of the domain gap. For the category-level alignment across different domains, we present an Inter-class Separation and Intra-class Aggregation (ISIA) mechanism. The key idea of the proposed ISIA is to close the feature distribution distance between the same categories and extend the feature distribution distance among different categories in the source and target domains. 

Firstly, we feed $\hat{x}^s\in \hat{X}^S$ and $x^t\in X^T$ into a shared encoder $F$ and two individual decoders $\{D^S, D^T\}$ to capture the features as:
\begin{equation}
	\begin{split}
		&f^s(\hat{x}^s), p^s(\hat{x}^s) = D^S(F(\hat{x}^s)) \\
		&f^t(x^t), p^t(x^t) = D^T(F(x^t))
	\end{split}
\end{equation}
where $\hat{x}^s$ represents the style-transferred source domain image as introduced in Sec.~\ref{sec-4.1}. $f^s(\hat{x}^s), f^t(x^t)\in \mathbb{R}^{D\times H\times W}$ are the semantic features with dimension $D$, $p^s(\hat{x}^s), p^t(x^t)\in \mathbb{R}^{N\times H\times W}$ are the probability predictions. In our implementation, $D$ is set to 2048 and $N_c$ represents the number of semantic categories.  For our category-level domain adaptation, the key is to align the same category and differentiate the different categories. In high dimensional space, features are sparsely distributed. We extract $\{c_i^s, c_i^t\ | c_i \in \mathbb{R}^{1\times N{_c}}, i=1,2,...,N\}$ from $\{p^s(\hat{x}^s), p^t(x^t)\}$ by selecting the corresponding channel. Thus for features those belong to the same category, our goal is to close the distance between source-domain features and target-domain features. For features those belong to different categories, the goal is to separate the feature distributions. We use cosine distance to measure the feature similarity of different categories:
\begin{equation}
	D_{cosine}(c_i, c_j) = \frac{{c_i}\cdot c_j}{||c_i||\times ||c_j||}, where    \quad i\neq j
\end{equation}
where $c_i$ and $c_j$ represent feature vector belonging to $i$-th and $j$-th class, respectively. Because the cosine distance ranges from -1 to 1, here we design Eqn.~(\ref{Eqn-Dsim}) to normalize the distance value to [0, 1] for training convenience.
\begin{equation}
	D_{sim}(c_i, c_j)=0.5+0.5\times D_{cosine}(c_i, c_j)
	\label{Eqn-Dsim}
\end{equation}
Here for all categories across domains, we pull closer features those belonging to the same category  and push away those belonging to different categories. Specifically, we use the L1 norm and the cosine similarity defined in Eqn.~(\ref{Eqn-Dsim}) to measure the embedding distance between the same and different categories, respectively. The inter-class separation and intra-class aggregation loss is defined as:
\begin{equation}
	\mathcal{L}_{ISIA} =\sum_{i=1}^N||c_i^s-c_i^t||_1 + \beta \sum_{i=1}^{N_c} \sum_{k=1, k\neq i}^{N_c} D_{sim}(c_i^s, c_k^t)
	\label{Eqn-Lmr}
\end{equation}
where $c_i^s$ and $c_i^t$ represent the feature of the $i$-th class of the input image belongs to the source domain and the target domain, respectively. $\beta$ is used to weigh the contribution of inter-class separation during the training. 

\subsection{Category-guided Instance level Alignment}
\label{Sec-AIM}
\cite{UDADT} splits the objects into background stuff that usually shares similar appearance across different domains, and foreground things that often have much larger variance across images. It indicates that the foreground classes may contribute the most discrepancy across different domains. Motivated by this observation, we focus on the foreground classes those have large appearance variation and design an Adaptive-weighted Instance Matching (AIM) strategy. However, due to the lack of instance-level annotations from the source domain, we first generate the instance masks by finding the disconnected regions for each class in the label map $L$ follows~\cite{UDADT}. By coarsely segmenting the intra-class semantic regions into multiple instances, the instance-level feature representations in one image is expressed as follows:
\begin{equation}
	\begin{split}
		&R_k =\{r_{k_1}, r_{k_2}, ..., r_{k_n}\} = \Gamma (L, k)  \\
		&\mathfrak{L}(r, f) = \frac{\sum_{(h, w)}r^{(h, w)}f^{(h, w)}}{max({\epsilon, \sum_{(h, w)}r^{(h, w)}})}
	\end{split}
	\label{Eqn-Rk}
\end{equation}
where $r_{k_i}$ represents the $i$-th ($i\in \{1,...,n\}$) binary mask of the connected region belonging to class $k$. $\Gamma$ is the operation to find the disconnected regions of class $k$ from the label mask $L$. $f$ is the feature map generated by the feature extractor network. $h$ and $w$ are the height and width of the feature maps. $\epsilon$ is a regularizing term. $\mathfrak{L}$ is the operation to generate the instance-level feature representation. 

Considering the category-level alignment described in Sec.~\ref{sec-ICSA}, we build a ranking list to measure the complexity of the category-level adaptation across domains. We denote category-level adaptation complexity for each class as $R_{ac}=\{\zeta_k|k=1,2, ... , N_{ins} \}$, where $N_{ins}$ is the category number of instance. $\zeta_k$ is computed by Eqn.~(\ref{Eqn-lambda}).
\begin{equation}
	\begin{split}
		\zeta_k& = \frac{||c_k^s-c_k^t||_1}{max(||c_i^s-c_i^t||_1)-min(||c_i^s-c_i^t||_1)} \\
		\eta_k &= \frac{\zeta_k}{max(||\zeta_i - \zeta_j||_1)}, i, j = 1, 2, ..., N_{ins}
	\end{split}
	\label{Eqn-lambda}
\end{equation}
where $k, i \in \{1,...,N_{ins}\}$. $\zeta_k$ is updated by every batch and $\eta_k$ is to avoid the weight saltus during the training. Thus the instance features across the source and target domains can be pulled closer by minimizing the cross-domain instance matching loss:
\begin{equation}
	\mathcal{L}_{AIM}=\sum_{i} \sum_{k\in N_{ins}} \frac{\eta_k}{|R_k^t|} \sum_{r^t \in R_k^t} \underset{j}{min}||\mathfrak{L}(r^t, f_i^t)-s_j^k||_1
	\label{Eqn-Lintra}
\end{equation}
where $i\in \{1,2,..., |X^T|\}$ and $R_k^t=\Gamma (L_{P_i}^t, k)$.  $s_j^k$ represents the $j$-th source domain semantic feature sample of class $k$. Here $\eta_k$ is used to weigh the instance-level alignment of $k$-th class.

\subsection{Integrated Objective}
We train our model in a two-step way. Firstly, due to the lack of the target domain labels, we train our model with an initial step defined in Eqn.~(\ref{Eqn-Initstep}). 
\begin{equation}
	\begin{split}
		\mathcal{L}_{init}& = \underset{G}{min}(\lambda_{seg}\mathcal{L}_{seg}^{S}+\lambda_{adv}\mathcal{L}_{adv}^f \\
		&+\lambda_{ISIA}\mathcal{L}_{ISIA}+\lambda_{AIM}\mathcal{L}_{AIM})+\underset{D}{min}\lambda_{D}\mathcal{L}_D
	\end{split}
	\label{Eqn-Initstep}
\end{equation}
Then we use self-supervised learning approach same to~\cite{UDADT} to generate pseudo labels to the pixels with high confidence of the predicted labels in the target domain training set images. Finally, we retrain our proposed models as follows:
\begin{equation}
	\begin{split}
		\mathcal{L}_{total}& = \underset{G}{min}(\lambda_{seg}(\mathcal{L}_{seg}^{S}+\mathcal{L}_{seg}^{T})+\lambda_{adv}\mathcal{L}_{adv}^f \\
		&+\lambda_{ISIA}\mathcal{L}_{ISIA}+\lambda_{AIM}\mathcal{L}_{AIM})+\underset{D}{min}\lambda_{D}\mathcal{L}_D
	\end{split}
	\label{Eqn-IO}
\end{equation}
where $\mathcal{L}_{seg}^S$ and $\mathcal{L}_{seg}^{T}$ are cross-entropy losses defined in Eqn.~(\ref{Eqn1}), which are used for measuring the prediction map of source domain and the target domain, respectively. $\lambda_{seg}, \lambda_{adv}$, $\lambda_{ISIA}$, $\lambda_{AIM}$ and $\lambda_D$ are the weight parameters for the losses.
The pseudocode of the proposed method is shown in Algorithm~\ref{alg-Pseudocode}.
\begin{algorithm}[htb]  
	\caption{Pseudocode of the proposed framework.} 
	\label{alg-Pseudocode}  
	\begin{algorithmic}[1]  
		\Require  
		The source domain images and labels $\{x^s, y^s|x^s\in X^S, y^s\in Y^S\}$, the target domain images $x^t\in X^T$ ; 
		\Ensure  
		Pixel-level prediction $\hat{y}_t$ of target domain images;
		\State Suppose: segmentation network \textbf{\emph{Seg}}, \emph{init\_iters=40k}, \emph{total\_iters=120k}, adapted model $\boldsymbol{M_{step1}}$, $\boldsymbol{M_{step2}}$;
		\State $\hat{x}^s\leftarrow CycleGAN(x^s)$, pair \{{$\hat{x}^s$, $y^s$}\};
		\State \textbf{for} \emph{curr\_iter} \textbf{in} \emph{init\_iters}:  
		\State 	\qquad $\hat{y}^s_{pred}\leftarrow$\emph{\textbf{Seg}(x$^{s}$)}, calculate $\mathcal{L}_{seg}^S$ \{forward pass\}
		\State  \qquad $f^s(\hat{x}^s), p^s(\hat{x}^s)\leftarrow D^{S}(F(\hat{x}^s))$ \{forward pass\}
		\State \qquad $f^t(\hat{x}^t), p^t(\hat{x}^t)\leftarrow D^{T}(F(\hat{x}^t))$ \{forward pass\}
		\State \qquad Calculate $\mathcal{L}_{adv}^{f}, \mathcal{L}_{D}$, $\mathcal{L}_{ISIA}, \mathcal{L}_{AIM}$
		\State \qquad Optimize $\mathcal{L}_{init}$ \{backward pass\}
		\State \textbf{return} $\boldsymbol{M_{step1}}$
		\State Generate pseudo label $\tilde{y}^t\in Y^T$ via $\boldsymbol{M_{step1}}$
		\State \textbf{for} $x_i^t$ in $X^T$:
		\State \qquad $\tilde{y}_i^t \leftarrow \boldsymbol{M_{step1}}(x_i^t)$  \{forward pass\}
		\State \textbf{for} \emph{curr\_iter} \textbf{in} \emph{total\_iters-init\_iters}: 
		\State \qquad  $\hat{y}^s_{pred}\leftarrow$\emph{\textbf{Seg}(x$^{s}$)},  $\hat{y}^t_{pred}\leftarrow$\emph{\textbf{Seg}(x$^{t}$)} \{forward pass\}
		\State \qquad Calculate $\mathcal{L}_{seg}^S, \mathcal{L}_{seg}^T$
		\State \qquad Repeat step 5-7
		\State \qquad Optimize $~\mathcal{L}_{total}$ \{backward pass\} 
		\State \textbf{return} $\boldsymbol{M_{step2}}$
		%\Return $\hat{y}^t$, $M_{ada}$;  
	\end{algorithmic}  
\end{algorithm}

\subsection{Network Architecture and Implementation}
For feature extractor, we directly utilize the DeepLab-v2~\cite{DeepLabv2} framework with ResNet-101~\cite{ResNet} pretrained on ImageNet~\cite{ImageNet} with 5 convolutional layers as the segmentation network. 
For discriminator network $D$, we adopt a similar structure with~\cite{CLAN}, which consists of 5 convolution layers with kernal 4$\times$4 with channel numbers $\{64, 128, 256, 512, 1\}$ and stride of 2. Each convolution layer is followed by a Leaky-ReLU~\cite{LeakyReLU} parameterized by $0.2$ negative slope between adjacent convolutional layers. The discriminator is implemented on the upsampled softmax output of the ASPP head. To train the segmentation network, we use SGD~\cite{SGD} as the optimizer for $G$ with a momentum of $0.9$, while using Adam~\cite{Adam} to optimize $D$ with $\beta_1=0.9$, $\beta_2=0.99$. Both optimizers are set a weight decay of $5\times 10^{-4}$. For SGD, the initial learning rate is set to $2.5\times 10^{-4}$ and decayed by a poly learning rate policy. For Adam, we initialize the learning rate to a fixed $5\times10^{-5}$ . In the first training stage, the network is trained for $40k$ iterations by optimizing Eqn.~(\ref{Eqn-Initstep}). After that we further optimize Eqn.~(\ref{Eqn-IO}) for a total of $120k$ iterations. We set $\lambda_{seg}=1$, $\lambda_{D}=1$, $\lambda_{adv}=0.001$ and batchsize as 1.  All experiments are conducted on a workstation with 4 NVIDIA 2080Ti GPU cards under CUDA 11.0.

\section{Experiments}
\subsection{Datasets}
\subsubsection{Street scenes}
Cityscapes~\cite{Cityscapes} is a real-world dataset with 5000 street scenes of resolution 2048$\times$1024. The dataset is split into training, validation and testing sets with 2975, 500, 1525 images, respectively. Following previous works~\cite{UDADT, CLAN}, we evaluate the models on the validation set. The Cityscapes images are resized to 1024$\times$512 for both the training and testing stage. The GTA5~\cite{GTA5} dataset consists of 24966 fine annotated synthetic images of resolution 1914$\times$1052. All the images are captured from the Grand Theft Auto V. And it shares all 19 classes with Cityscapes. 
%For comparison fairness, we follow~\cite{AdaptSegNet} and resize GTA5 images to 1280$\times$720 resolution. 
SYNTHIA~\cite{SYNTHIA} is another synthetic image dataset that contains 9400 images of resolution 1280$\times$760.  Similar to~\cite{CLAN, SSFDAN, UDADT}, the models are evaluated on Cityscapes validation set for the 13 common classes between SYNTHIA and Cityscapes. 
\subsubsection{Remote sensing images}
\begin{figure*}[htb]
	\centering
	\includegraphics[scale=0.5]{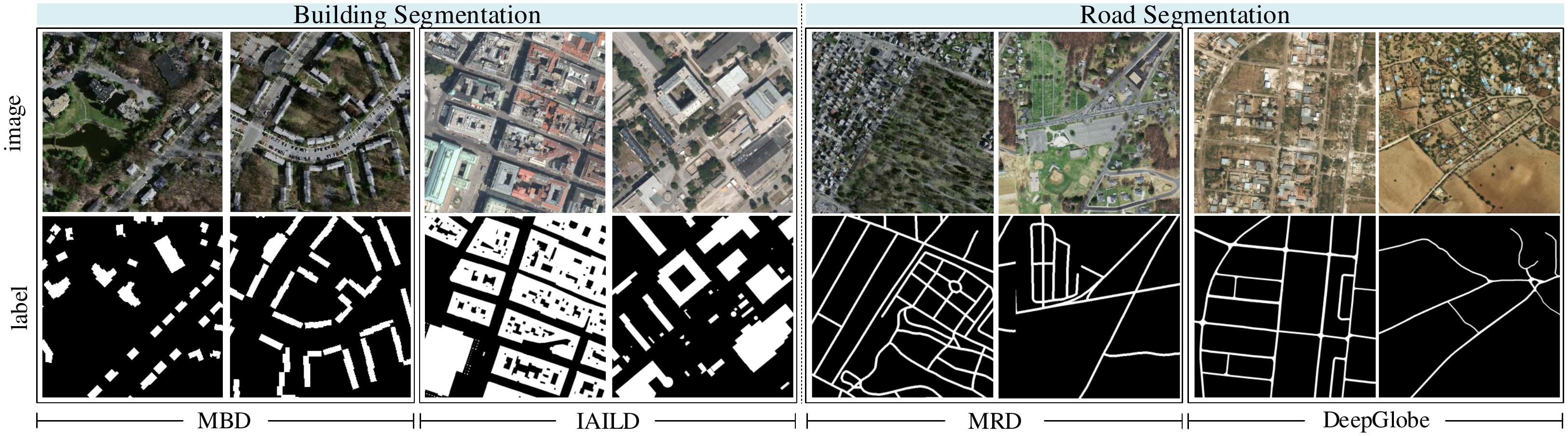}
	\caption{The qualitative comparison of cross-domain building segmentation datasets and cross-domain road segmentation datasets. }
	\label{fig-RSI_dataset}
\end{figure*}

Domain adaptation provides a way of using the existing labeled data to run inference in unlabeled data in remote sensing image interpretation. We organize two cross-domain semantic segmentation datasets for building segmentation and road segmentation on the basis of public data, respectively. Inria Aerial Image Labeling Dataset (IAILD)~\cite{IAILD} is a large-scale dataset for building extraction with a spatial resolution of 0.3 m and 180 labeled images with 5000$\times$5000 pixels, covering different urban
areas and the same areas in different time period. Massachusetts Building Dataset (MBD)~\cite{MnihThesis} contains 151 sets of aerial images and corresponding single-channel label images with 2 classes.  For training convenience, we randomly cut the image into $512\times 512$ patches. Massachusetts Road Dataset (MRD)~\cite{MnihThesis} consists of 1171 aerial images and corresponding binary label maps, each image is 1500$\times$1500 pixels in size with a spatial resolution of 1 m , covering an area of 2.25 km$^{2}$. DeepGlobe~\cite{DeepGlobe} for road extraction contains 850 images with 2 classes annotations with size of 1024$\times$1024 and the ground resolution of the image pixels is 0.5m/pixel. We also cut the images into 512$\times$512 patches due to the GPU memory limitation. The cross domain datasets have difference in imaging area, object gray scale, object appearance, image annotation format, spatial resolution, etc. The datasets details are shown in Table~\ref{table-RSI-dataset}. Fig.~\ref{fig-RSI_dataset} shows the qualitative comparison between the different domains.
\begin{table*}[htbp]
\centering
\small
\caption{Cross-domain semantic segmentation datasets.}
\setlength{\tabcolsep}{1mm}{
	{\begin{tabular*}{0.8\textwidth}{@{\extracolsep{\fill}}c|ccccc@{}}
	\toprule[0.4mm]
	\midrule
	Type&Task&Shared classes&Spatial-resolution&Train set&Val set \\
	\midrule
	\multirow{2}{*}{\makecell{Street\\scenes}}&GTA5$\rightarrow$Cityscapes&19&-&24966&500 \\
	\cline{2-6}
	&SYNTHIA$\rightarrow$Cityscapes&13&-&9400&500 \\
	\midrule 
	\multirow{4}{*}{\makecell{Remote\\sensing\\images}}&MBD$\rightarrow$IAILD&2&1.0m$\rightarrow$0.3m&4110&800 \\
	\cline{2-6}
	&IAILD$\rightarrow$MBD&2&0.3m$\rightarrow$1.0m&2800&350 \\
	\cline{2-6}
	&MRD$\rightarrow$DeepGlobe&2&1.0m$\rightarrow$0.5m&4388&350 \\
	\cline{2-6}
	&DeepGlobe$\rightarrow$MRD&2&0.5m$\rightarrow$1.0m&500&567 \\
	\midrule
	\bottomrule[0.4mm]
	\end{tabular*}}{}}	
	\label{table-RSI-dataset}
\end{table*}

We compute PASCAL VOC intersection-over-union (IoU)~\cite{PASCAL_VOC} for evaluation: 
\begin{equation}
	IoU = \frac{TP}{TP+FP+FN}
\end{equation}
where TP, FP and FN are the number of true positive, false positive and false negative pixels, respectively.

\begin{table*}[htbp]
	\centering
	\footnotesize
	\caption{Quantitative comparison on "GTA5$\rightarrow$Cityscapes" in terms of per-class IoUs and mIoU (\%). All the results are generated from the ResNet-101-based models. The first and second best results are highlighted in red and blue, respectively.}
	\setlength{\tabcolsep}{1mm}{
		{\begin{tabular*}{1.0\textwidth}{@{\extracolsep{\fill}}l|ccccccccccccccccccc|c@{}}
				\toprule[0.4mm]
				\hline
				Method&\begin{sideways}road\end{sideways}&\begin{sideways}sidewalk\end{sideways}&\begin{sideways}building\end{sideways}&\begin{sideways}wall\end{sideways}&\begin{sideways}fence\end{sideways}&\begin{sideways}pole\end{sideways}&\begin{sideways}light\end{sideways}&\begin{sideways}sign\end{sideways}&\begin{sideways}vege.\end{sideways}&\begin{sideways}terrain\end{sideways}&\begin{sideways}sky\end{sideways}&\begin{sideways}person\end{sideways}&\begin{sideways}rider\end{sideways}&\begin{sideways}car\end{sideways}&\begin{sideways}truck\end{sideways}&\begin{sideways}bus\end{sideways}&\begin{sideways}train\end{sideways}&\begin{sideways}motor\end{sideways}&\begin{sideways}bicycle\end{sideways}&\begin{sideways}mIoU\end{sideways} \\
				\midrule
				AdaptSeg~\cite{AdaptSegNet}&86.5&36.0&79.9&23.4&23.3&23.9&35.2&14.8&83.4&33.3&75.6&58.5&27.6&73.7&32.5&35.4&3.9&30.1&28.1&42.4 \\
				CBST~\cite{CBST}&89.6&\textcolor{red}{58.9}&78.5&33.0&22.3&\textcolor{red}{41.4}&\textcolor{red}{48.2}&\textcolor{blue}{39.2}&83.6&24.3&65.4&49.3&20.2&83.3&\textcolor{blue}{39.0}&\textcolor{blue}{48.6}&12.5&20.3&35.3&47.0 \\
				CLAN~\cite{CLAN}&87.0&27.1&79.6&27.3&23.3&28.3&35.5&24.2&83.6&27.4&74.2&58.6&28.0&76.2&33.1&36.7&6.7&31.9&31.4&43.2 \\
				SIBAN~\cite{SIBAN} &88.5 &35.4 &79.5 &26.3 &24.3 &28.5 &32.5 &18.3 &81.2 &40.0 &76.5& 58.1& 25.8 &82.6 &30.3 &34.4 &3.4 &21.6 &21.5 &42.6 \\ 
				MaxSquare~\cite{MaxSquare} &88.1&27.7&80.8&28.7&19.8&24.9&34.0&17.8&83.6&34.7&76.0&58.6&28.6&84.1&37.8&43.1&7.2&32.2&34.2&44.3 \\
				AdvEnt~\cite{ADVENT}&89.4&33.1&81.0&26.6&26.8&27.2&33.5&24.7&83.9&36.7&78.8&58.7&30.5&84.8&38.5&44.5&1.7&31.6&32.4&45.5 \\
				DPR~\cite{DPR}&\textcolor{blue}{92.3}&\textcolor{blue}{51.9}&82.1&29.2&25.1&24.5&33.8&33.0&82.4&32.8&82.2&58.6&27.2&84.3&33.4&46.3&2.2&29.5&32.3&46.5 \\
				PyCDA~\cite{PyCDA}&90.5&36.3&84.4&32.4&28.7&34.6&36.4&31.5&\textcolor{red}{86.8}&37.9&78.5&62.3&21.5&85.6&27.9&34.8&\textcolor{red}{18.0}&22.9&\textcolor{red}{49.3}&47.4 \\
				SSF-DAN~\cite{SSFDAN}&90.3&38.9&81.7&24.8&22.9&30.5&37.0&21.2&84.8&38.8&76.9&58.8&30.7&85.7&30.6&38.1&5.9&28.3&36.9&45.4 \\
				DISE~\cite{DISE}&91.5 &47.5 &82.5 &31.3 &25.6 &33.0 &33.7 &25.8 &82.7 &28.8 &82.7 &\textcolor{blue}{62.4} &30.8 &85.2 &27.7 &34.5 &6.4 &25.2 &24.4 &45.4  \\
				DLOW~\cite{DLOW}&87.1 &33.5 &80.5 &24.5 &13.2 &29.8 &29.5 &26.6 &82.6 &26.7 &81.8 &55.9 &25.3 &78.0 &33.5 &38.7 &0.0 &22.9 &34.5 &42.3  \\
				FADA~\cite{FADA}&\textcolor{red}{92.5}&47.5&\textcolor{blue}{85.1}&\textcolor{blue}{37.6}&\textcolor{red}{32.8}&33.4&33.8&18.4&85.3&37.7&83.5&\textcolor{red}{63.2}&\textcolor{red}{39.7}&\textcolor{red}{87.5}&32.9&47.8&1.6&\textcolor{red}{34.9}&\textcolor{blue}{39.5}&\textcolor{blue}{49.2} \\
				IntraDA~\cite{IntraDA}&90.6&37.1&82.6&30.1&19.1&29.5&32.4&20.6&\textcolor{blue}{85.7}&\textcolor{blue}{40.5}&79.7&58.7&31.1&\textcolor{blue}{86.3}&31.5&48.3&0.0&30.2 &35.8&46.3 \\
				Wang et al.~\cite{UDADT}&90.6&44.7&84.8&34.3&28.7&31.6&35.0&37.6&84.7&\textcolor{red}{43.3}&\textcolor{blue}{85.3}&57.0&31.5&83.8&\textcolor{red}{42.6}&48.5&1.9&30.4&39.0&\textcolor{blue}{49.2} \\
				ASA~\cite{ASA}&89.2&27.8&81.3&25.3&22.7&28.7&36.5&19.6&83.8&31.4&77.1&59.2&29.8&84.3&33.2&45.6&\textcolor{blue}{16.9}&\textcolor{blue}{34.5}&30.8&45.1  \\
				\textbf{Ours}&91.8&48.7&\textcolor{red}{85.6}&\textcolor{red}{38.1}&\textcolor{blue}{31.8}&\textcolor{blue}{35.7}&\textcolor{blue}{39.5}&\textcolor{red}{40.3}&85.3&\textcolor{blue}{40.5}&\textcolor{red}{85.9}&62.2&\textcolor{blue}{32.3}&84.2&31.4&\textcolor{red}{52.2}&9.9&31.0&36.1&\textcolor{red}{50.7} \\
				\hline
				\bottomrule[0.4mm]
		\end{tabular*}}{}}	
	\label{table-GTA2Cityscapes}
\end{table*}
\begin{table*}[htbp]
	\centering
	\footnotesize
	\caption{Quantitative comparison on "SYNTHIA$\rightarrow$Cityscapes" in terms of per-class IoUs and mIoU (\%). All the results are generated from the ResNet-101-based models. The mIoU column donated the mean IoU over 13 categories shared by the SYNTHIA and Cityscapes. The first and second best results are highlighted in red and blue, respectively.}
	\setlength{\tabcolsep}{3.1mm}{
		{\begin{tabular*}{1.0\textwidth}{l|ccccccccccccc|c}
				\toprule[0.4mm]
				\hline
				Method&\begin{sideways}road\end{sideways}&\begin{sideways}sidewalk\end{sideways}&\begin{sideways}building\end{sideways}&\begin{sideways}light\end{sideways}&\begin{sideways}sign\end{sideways}&\begin{sideways}vege.\end{sideways}&\begin{sideways}sky\end{sideways}&\begin{sideways}person\end{sideways}&\begin{sideways}rider\end{sideways}&\begin{sideways}car\end{sideways}&\begin{sideways}bus\end{sideways}&\begin{sideways}motor\end{sideways}&\begin{sideways}bicycle\end{sideways}&\begin{sideways}mIoU\end{sideways} \\
				\midrule 
				AdaptSeg~\cite{AdaptSegNet}&84.3&42.7&77.5&4.7&7.0&77.9&82.5&54.3&21.0&72.3&32.2&18.9&32.3&46.7 \\
				CLAN~\cite{CLAN}&81.3&37.0&80.1&16.1&13.7&78.2&81.5&53.4&21.2&73.0&32.9&22.6&30.7&47.8 \\
				MaxSquare~\cite{MaxSquare} &77.4&34.0&78.7&5.8&9.8&80.7&83.2&\textcolor{blue}{58.5}&20.5&74.1&32.1&11.0&29.9&45.8\\
				AdvEnt~\cite{ADVENT}&\textcolor{blue}{85.6}&42.2&79.7&5.4&8.1&80.4&\textcolor{blue}{84.1}&57.9 &23.8&73.3&36.4&14.2&33.0&48.0 \\
				DPR~\cite{DPR}&82.4&38.0&78.6&3.9&11.1&75.5&\textcolor{red}{84.6}&53.5&21.6&71.4&32.6&19.3&31.7&46.5 \\
				FADA~\cite{FADA}&84.5&40.1&\textcolor{red}{83.1}&\textcolor{blue}{20.1}&\textcolor{blue}{27.2}&\textcolor{red}{84.8}&84.0&53.5&22.6&\textcolor{red}{85.4}&\textcolor{red}{43.7}&\textcolor{blue}{26.8}&27.8&\textcolor{blue}{52.5} \\
				IntraDA~\cite{IntraDA}&84.3&37.7&79.5&9.2&8.4&80.0&\textcolor{blue}{84.1}&57.2&23.0&78.0&38.1&20.3&36.5&48.9 \\
				Wang et al.~\cite{UDADT}&83.0&\textcolor{blue}{44.0}&80.3&17.1&15.8&80.5&81.8&\textcolor{red}{59.9}&\textcolor{red}{33.1}&70.2&37.3&\textcolor{red}{28.5}&\textcolor{blue}{45.8}&52.1 \\
				ASA~\cite{ASA}&\textcolor{red}{91.2}&\textcolor{red}{48.5}&80.4&5.5&5.2&79.5&83.6&56.4&21.0&\textcolor{blue}{80.3}&36.2&20.0&32.9 &49.3\\
				\textbf{Ours}&78.9&35.7&\textcolor{blue}{81.3}&\textcolor{red}{26.4}&\textcolor{red}{31.5}&\textcolor{blue}{81.5}&83.5&53.4&\textcolor{blue}{26.1}&78.8&\textcolor{blue}{40.0}&\textcolor{red}{28.5}&\textcolor{red}{48.8}&\textcolor{red}{53.4}\\
				\hline
				\bottomrule[0.4mm]
		\end{tabular*}}{}}	
	\label{table-SYNTHIA2Cityscapes}
\end{table*}

\begin{figure*}
	\centering
	\includegraphics[scale=0.5]{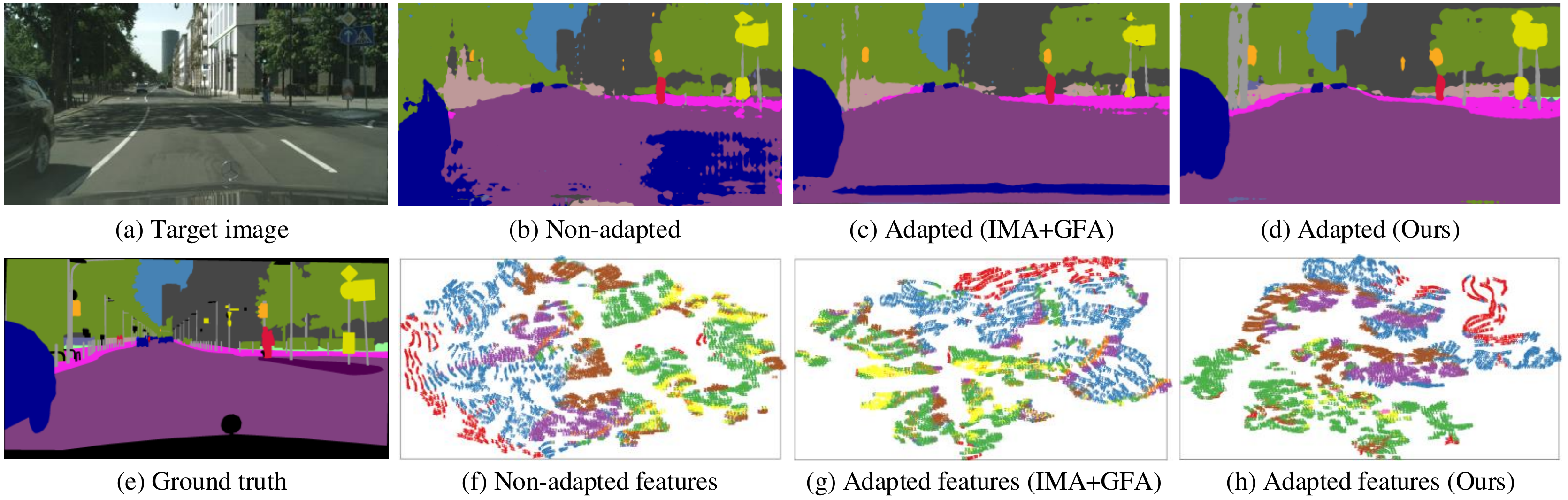}
	\caption{Contrastive analysis of the feature distributions. (a): A target image; (b): A segmentation map of the model trained on source domain dataset only. Although the segmentation result is poor, many classes can still be correctly segmented, which indicates some classes are originally aligned without any adaptation.  (c): Adapted segmentation map by adopting IMA+GFA. The segmentation performance improvement is not obvious because the IMA focuses on the appearance transferring and GFA strategy uses a simple adversarial learning in global feature output space. They lack the attention on category confusion problem.  (d): Adapted result of our model. The pixels of confusable classes are well classified. Additionally, we use t-SNE~\cite{TSNE} to map the high-dimensional features of (b), (c), (d) to 2D space shown in (f), (g), (h), respectively.}
	\label{fig-TSNE_features}
\end{figure*}

\begin{figure*}[htbp]
	\centering
	\includegraphics[scale=0.42]{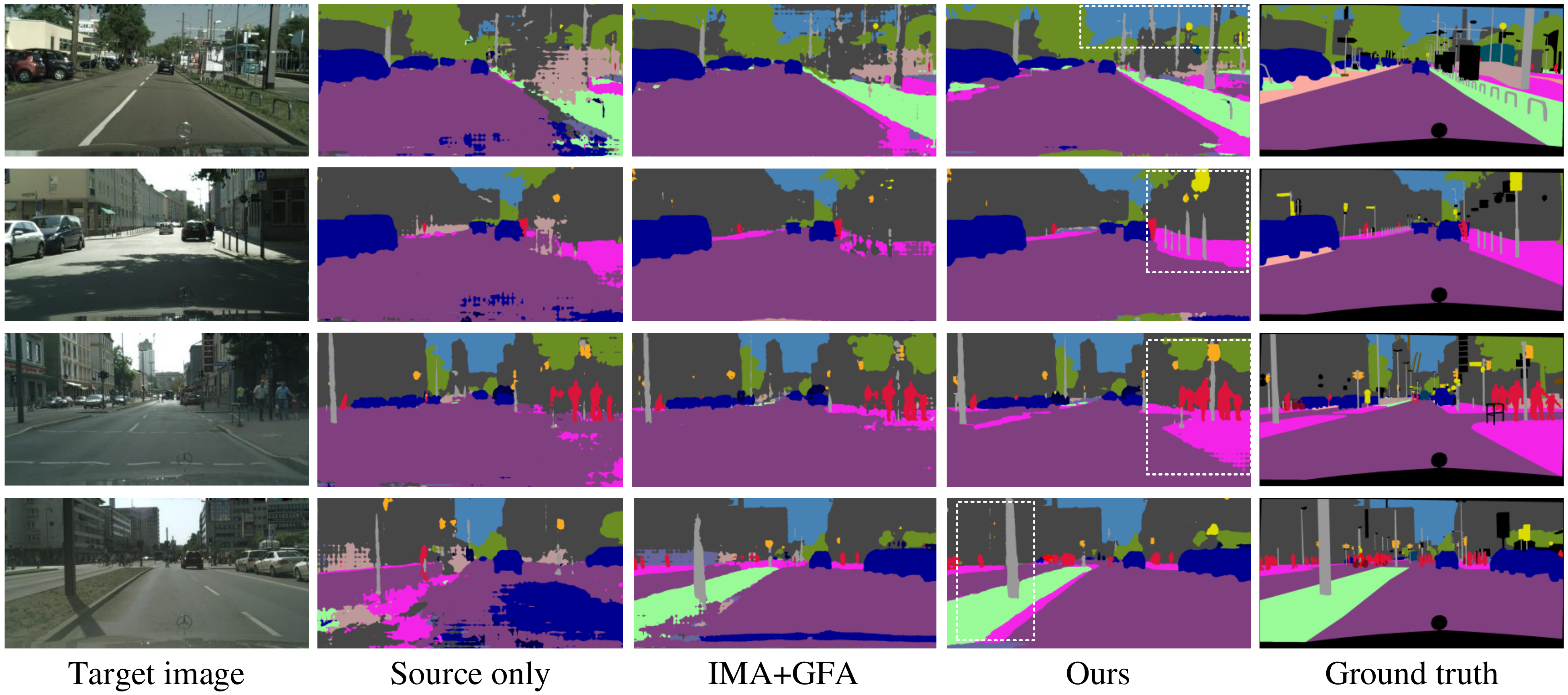}
	\caption{Qualitative visualizations from Cityscapes validation set. For each target image, we show the corresponding non-adapted (Source only) result, the adapted result with image-level adaptation and global feature-level adaptation (IMA+GFA), the adapted result produced by our proposed model and the ground truth.}
	\label{fig-visualization}
\end{figure*}
\subsection{Performance on Street Scenes}
\subsubsection{GTA5$\rightarrow$Cityscapes} 
\label{Sec-expG2C} 
\textbf{Overall results.} We compare the proposed model with the state-of-the-art UDA methods~\cite{AdaptSegNet, CLAN, SIBAN, ADVENT, DPR, PyCDA, SSFDAN, DISE, DLOW, CBST, ASA, FADA,  IntraDA, UDADT} in Table~\ref{table-GTA2Cityscapes}. Our method shows strong adaptation efficiency of the model in the target domain and achieves the highest IoU in five sub-categories and the second highest IoU in another five sub-categories, especially in confusable categories like \emph{building}, \emph{sign} and \emph{bus}, etc.. In terms of all categories, the proposed model achieves a new state-of-the-art performance with the mIoU of 50.7\%. 

\begin{figure}
	\includegraphics[scale=0.37]{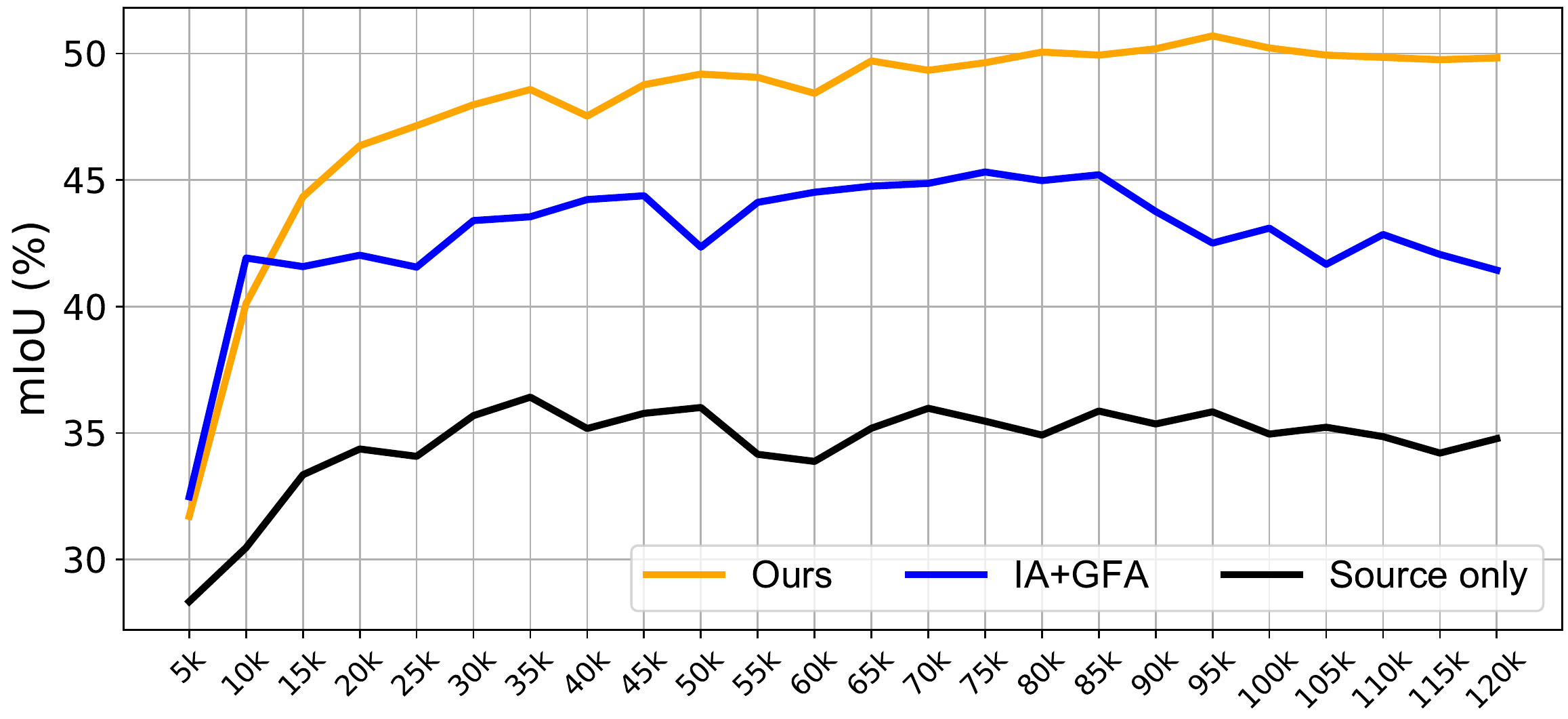}
	\caption{mIoU comparison on Cityscapes validation set. The model is tested every $5k$ iterations on GTA5$\rightarrow$Cityscapes. }
	\label{fig-plot_mIoU}
\end{figure}
\textbf{Module contributions.} We first assess the contribution of each module to the overall performance in Table~\ref{table-abla-G2C}. If the model is simply trained on the source domain dataset, it achieves an mIoU of 36.6\%. As introduced in Sec~\ref{sec-4.1}, we conduct image-level adaptation by transferring source image style to target domain~\cite{CycleGAN} and the model achieves 42.5\% mIoU. Through adversarial learning on the output space with adversarial loss proposed in~\cite{AdaptSegNet}, the mIoU is further improved to 45.3\%. The IMA and GFA strategies attempt to reduce domain shift in a holistic view but ignore semantic-level information. Then we employ the proposed ISIA to train the framework and set $\lambda_{ISIA}=0.001$ with the same weight of $\lambda_{adv}$,  the model achieves an mIoU of 49.6\%. Using the AIM module proposed in Sec.~\ref{Sec-AIM} and setting $\lambda_{AIM}=0.001$, the model achieves an mIoU of 50.7\% by optimizing Eqn~(\ref{Eqn-IO}). Same to~\cite{UDADT}, we split the objects of \emph{pole}, \emph{light}, \emph{sign}, \emph{person}, \emph{rider}, \emph{car}, \emph{truck}, \emph{bus}, \emph{train}, \emph{motor} and \emph{bike} into foreground classes and others into background classes. We focus on performing AIM on the foreground classes because they are hard to be aligned due to the large intra-class variance. Specifically, the background classes normally cover large areas and the features are easily to be distinguished. While the foreground classes usually have distinct variance in shape, texture and illuminance among instances so they are possibly to be misclassified. Fig.~\ref{fig-plot_mIoU} presents the mIoU variance comparison with the increase of iteration. The proposed method shows a steadier performance and achieves a large gain compared with the global feature-level adaptation approach~\cite{AdaptSegNet}. We further present a contrastive analysis for the feature distributions in Fig.~\ref{fig-TSNE_features}. Visually, the proposed model displays higher classification accuracy in the segmentation result in such a complex scene. And from the features distribution, the proposed method can enforce intra-class features closer and the inter-class features further apart. Together with the quantitative results in Fig.~\ref{fig-visualization}, the proposed method can effectively improve adaptation efficiency for each category and reduce pixels misclassification especially in complex scenes.

\textbf{Parameters study.} We show the influence of $\beta$ defined in Eqn.~(\ref{Eqn-Lmr})  to validate the contribution of inter-class separation and intra-class aggregation, respectively. As  shown in Table~\ref{table-abla-beta}, the model achieves the highest mIoU when $\beta=1.0$. Hence we argue that the weight of $\beta$ should not be either too large or too small. To our best knowledge, if $\beta$ is too small, the contribution of inter-class separation strategy is mild and there is high probability of pixels misclassification. While if $\beta$ is too large, it leads to a drop on the segmentation accuracy. Because the influence of inter-class separation portion is violent that may override the intra-class aggregation efficiency.  In our implementation, the best performance occurred when $\beta=1.0$, which is fixed for the following experiments.

Next we discuss the contribution of the proposed ISIA by adjusting its weight coefficient $\lambda_{ISIA}$ given $\lambda_{AIM}=0.001$. As shown in Table~\ref{table-abla-ISIA}, when $\lambda_{ISIA}=0.001$, which equals to $\lambda_{adv}$, the model achieves the highest mIoU. From the experimental results, a small $\lambda_{ISIA}$ may have little improvement on reducing the domain shift. A large $\lambda_{ISIA}$ tends to pull the features those have large intra-class variance too much closer to the same feature sample and even aggravate the pixels misclassification, which leads to segmentation accuracy decline. By setting $\lambda_{ISIA}=0.001$,  the influence of $\lambda_{AIM}$ is also explored in Table~\ref{table-abla-AIM}. We follow~\cite{UDADT} to adapt 10 instance features at maximum for each class from the target domain to the source domain. Our model achieves the best performance when $\lambda_{AIM}=0.001$. If $\lambda_{AIM}$ is too small, the proposed instance-level alignment can bring a limited improvement to the model. On the other hand, if $\lambda_{AIM}$ is too large, it could worsen the adaptation performance. This is because the instance features of small regions may be mixed with noisy regions due to the bottleneck of the segmentation model.

% ablation study on GTA5 to Cityscapes
\begin{table}[htbp]
	\centering
	\small
	\caption{Ablation study on GTA5$\rightarrow$Cityscapes. IMA donates the image-level adaptation; GFA stands for global feature-level adaptation; ISIA is the proposed inter-class separation and intra-class aggregation mechanism. AIM indicates the proposed adaptive-weighted instance matching strategy.}
	{\begin{tabular}{l|cccc|c}
			\toprule[0.4mm]
			%\midrule
			Method&IMA&GFA&ISIA&AIM&mIoU(\%) \\
			\midrule
			Source only&&&&&36.6 \\
			\midrule
			+IMA~\cite{BidirectionalLF}&\checkmark&&&&42.5 \\
			+GFA\cite{AdaptSegNet}&\checkmark&\checkmark&&&45.3 \\
			+ISIA&\checkmark&\checkmark&\checkmark&&49.6 \\
			+AIM&\checkmark&\checkmark&&\checkmark&48.8 \\
			+\emph{all}&\checkmark&\checkmark&\checkmark&\checkmark&\textbf{50.7} \\
			\midrule
			Target only&&&&&65.1 \\
			%\hline
			\bottomrule[0.4mm]
	\end{tabular}}{}
	\label{table-abla-G2C}
\end{table}

\begin{table}[htbp]
	\centering
	\small
	\caption{Ablation study on SYNTHIA$\rightarrow$Cityscapes. IMA donates the image-level adaptation; GFA stands for global feature-level adaptation; ISIA is the proposed inter-class separation and intra-class aggregation mechanism. AIM indicates the proposed adaptive-weighted instance matching strategy.}
	{\begin{tabular}{l|cccc|c}
			\toprule[0.4mm]
			Method&IMA&GFA&ISIA&AIM&mIoU(\%) \\
			\midrule
			Source only&&&&&38.6 \\
			\midrule
			+IMA~\cite{BidirectionalLF}&\checkmark&&&&42.4 \\
			+GFA~\cite{AdaptSegNet}&\checkmark&\checkmark&&&45.6 \\
			+ISIA&\checkmark&\checkmark&\checkmark&&52.5 \\
			+AIM&\checkmark&\checkmark&&\checkmark&51.4	 \\
			+\emph{all}&\checkmark&\checkmark&\checkmark&\checkmark&\textbf{53.4 }\\
			\midrule
			Target only&&&&&71.7 \\
			\bottomrule[0.4mm]
	\end{tabular}}{}
	\label{table-abla-S2C}
\end{table}

\begin{table}[h]
	\centering
	\small
	\caption{Influence of $\beta$ defined in Eqn~(\ref{Eqn-Lmr}) on GTA5$\rightarrow$Cityscapes.}
	\setlength{\tabcolsep}{1.5mm}{\begin{tabular}{c|ccccc}
			\toprule[0.5mm]
			$\beta$&0.1&0.5&1.0&2.0&5.0\\
			\midrule
			mIoU(\%)&49.8&50.3&\textbf{50.7}&50.4&50.1\\
			\bottomrule[0.4mm]
	\end{tabular}}{}
	\label{table-abla-beta}
\end{table}

\begin{table}[h]
	\centering
	\small
	\caption{SENSITIVITY ANALYSIS of $\lambda_{ISIA}$ given $\lambda_{AIM}=0.001$.}
	\setlength{\tabcolsep}{0.8mm}{\begin{tabular}{c|cccccc}
			\toprule[0.5mm]
			\multicolumn{7}{c}{GTA5$\rightarrow$Cityscapes} \\
			\midrule
			$\lambda_{ISIA}$&0.0001&0.0005&0.001&0.005&0.01&0.02\\
			\midrule
			mIoU(\%)&49.2&49.6&\textbf{50.7}&50.2&50.0&49.3 \\
			\bottomrule[0.4mm]
	\end{tabular}}{}
	\label{table-abla-ISIA}
\end{table}

\begin{table}[!h]
	\centering
	\small
	\caption{SENSITIVITY ANALYSIS of $\lambda_{AIM}$ given $\lambda_{ISIA}=0.001$.}
	\setlength{\tabcolsep}{0.8mm}{\begin{tabular}{c|cccccc}
			\toprule[0.4mm]
			\multicolumn{7}{c}{GTA5$\rightarrow$Cityscapes} \\
			\midrule
			$\lambda_{AIM}$&0.0001&0.0005&0.001&0.005&0.01&0.02\\
			\midrule
			mIoU(\%)&50.2&50.4&\textbf{50.7}&50.4&50.1&50.0\\
			\bottomrule[0.4mm]
	\end{tabular}}{}
	\label{table-abla-AIM}
\end{table}

\subsubsection{SYNTHIA$\rightarrow$Cityscapes}
Following the same hyper parameters discussed in Sec.~\ref{Sec-expG2C}, we evaluate the proposed model on the SYNTHIA$\rightarrow$Cityscapes task compared with~\cite{AdaptSegNet, CLAN, MaxSquare, ADVENT, DPR, ASA, FADA, IntraDA, UDADT}. As shown in Table~\ref{table-SYNTHIA2Cityscapes}, our model achieves the highest mIoU with 53.4\% in terms of the performance on 13 common classes.

The contribution of each module is also analyzed on this adaptation task. From Table~\ref{table-abla-S2C}, the model achieves an mIoU of 38.6\% when it is trained on the source domain dataset only. The image-level adaptation brings 3.8\% mIoU improvement to 42.4\%. By using the GFA module, the mIoU is thereby improved to 45.6\%.  Here, the IA and GFA reach their performance bottleneck due to the large domain shift between the source domain and the target domain. By adding the ISIA module, the model achieves a large gain of segmentation accuracy to an mIoU of 52.5\%. Using the proposed AIM can further improve the mIoU to 53.4\%.  Experimental results prove the proposed method has a great impact on reducing the domain shift through the efficient category-level and instance-level alignments.

\subsection{Performance on Remote Sensing Images}
To extend the proposed model to more application fields, we carry our method on cross-domain remote sensing images on two tasks, i.e., cross-domain building segmentation and cross-domain road segmentation.
\subsubsection{Cross-domain building segmentation}
	\begin{figure}[t]
	\includegraphics[scale=0.54]{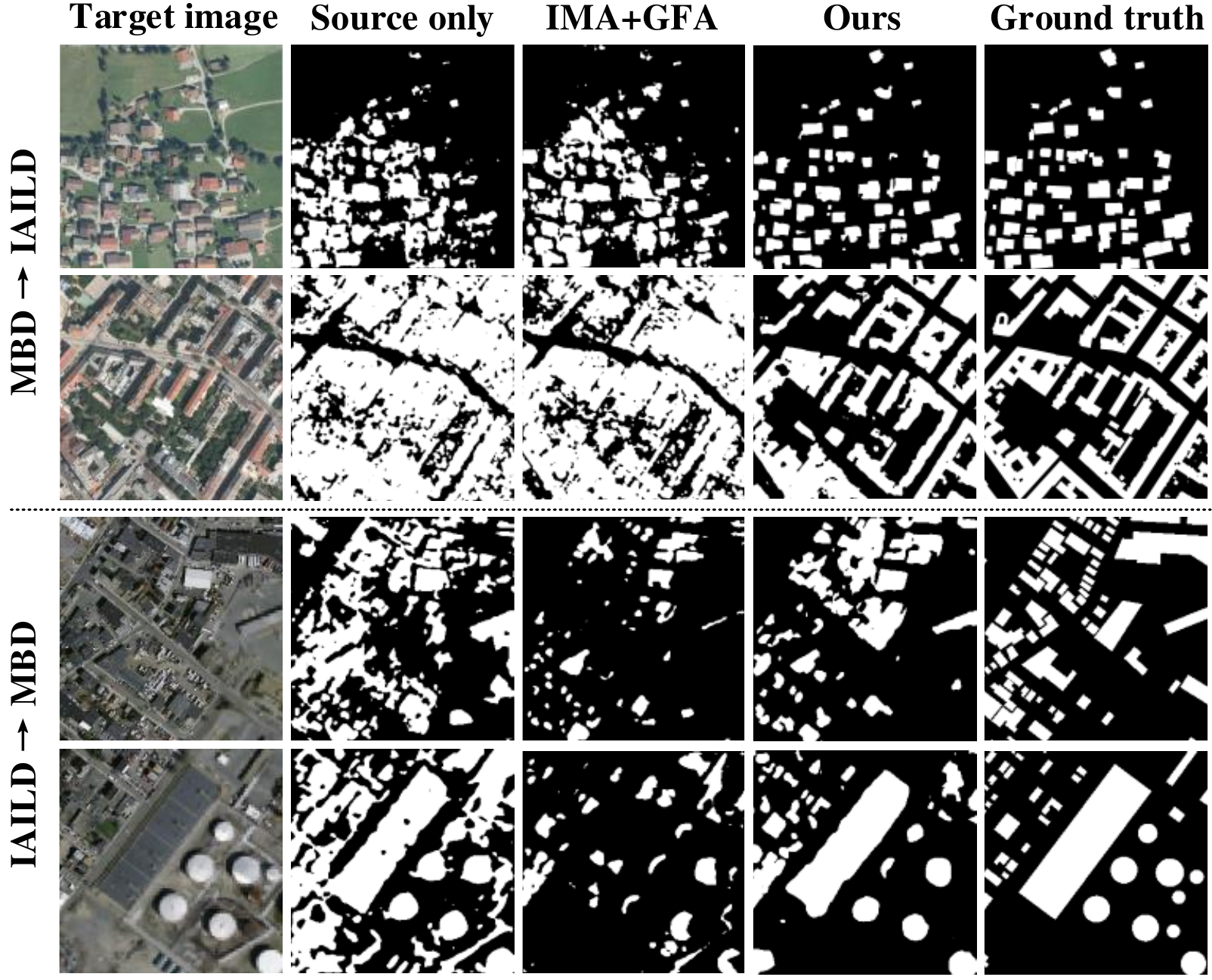}
	\caption{Qualitative visualizations on cross-domain building segmentation task. For each target image, we show the corresponding non-adapted (Source only) result, the adapted result with image-level adaptation and global feature-level adaptation (IMA+GFA), the adapted results produced by our proposed model and the ground truth.}
	\label{fig-visualization_building}		
\end{figure}
\begin{figure}[t]
	\includegraphics[scale=0.54]{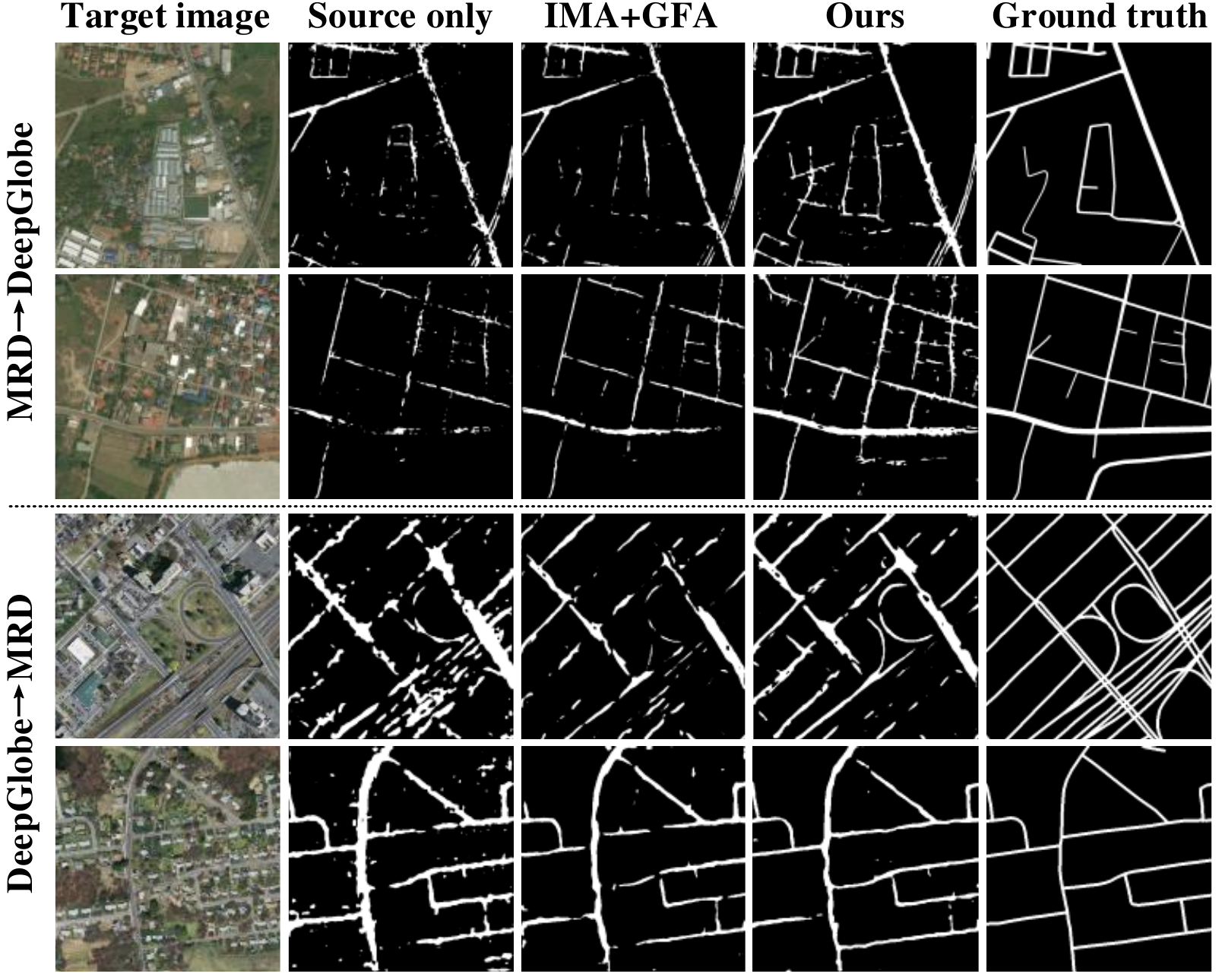}
	\caption{Qualitative visualizations on cross-domain road segmentation task. For each target image, we show the corresponding non-adapted (Source only) result, the adapted result with image-level adaptation and global feature-level adaptation (IMA+GFA), the adapted results produced by our proposed model and the ground truth.}
	\label{fig-visualization_road}		
\end{figure}

\begin{table}[ht]
	\centering
	\small
	\caption{Ablation study on cross-domain building segmentation task.  IMA donates the image-level adaptation; GFA stands for global feature-level adaptation; ISIA is the proposed inter-class separation and intra-class aggregation mechanism. AIM indicates the proposed adaptive-weighted instance matching strategy.}
			\setlength{\tabcolsep}{0.45mm}{
				{\begin{tabular*}{0.48\textwidth}{@{\extracolsep{\fill}}c|l|cccc|ccc@{}}
						\toprule[0.4mm]
						%\midrule
						Task&Method&IMA&GFA&ISIA&AIM&Build.&Bg.&mIoU(\%) \\
						\midrule
						\multirow{7}{*}{\makecell{MBD\\$\downarrow$\\IAILD}}&\emph{Source only}&&&&&67.4&84.8&76.1\\
						\cline{2-9}
						&+IMA~\cite{BidirectionalLF}&\checkmark&&&&71.6&88.6&80.1 \\
						&+GFA~\cite{AdaptSegNet}&\checkmark&\checkmark&&&72.9&88.9&80.9 \\
						&+ISIA&\checkmark&\checkmark&\checkmark&&73.3&89.2&81.3 \\
						&+AIM&\checkmark&\checkmark&&\checkmark&73.1&89.1&81.1	 \\
						&+\emph{all}&\checkmark&\checkmark&\checkmark&\checkmark&\textbf{73.9}&\textbf{89.6}&\textbf{81.7} \\
						\cline{2-9}
						&\emph{Target only}&&&&&75.1&90.2&82.6 \\
						%\hline
						\midrule
						\multirow{7}{*}{\makecell{IAILD\\$\downarrow$\\MBD}}&\emph{Source only}&&&&&35.8&87.9&61.8 \\
						\cline{2-9}
						&+IMA~\cite{BidirectionalLF}&\checkmark&&&&39.1&87.8&63.5 \\
						&+GFA~\cite{AdaptSegNet}&\checkmark&\checkmark&&&45.0&88.7&66.9 \\
						&+ISIA&\checkmark&\checkmark&\checkmark&&53.2&89.7&71.4 \\
						&+AIM&\checkmark&\checkmark&&\checkmark&52.7&89.3&71.0	 \\
						&+\emph{all}&\checkmark&\checkmark&\checkmark&\checkmark&\textbf{53.8}&\textbf{90.0}&\textbf{71.9} \\
						\cline{2-9}
						&\emph{Target only}&&&&&63.9&92.4&78.2 \\
						\bottomrule[0.4mm]
				\end{tabular*}}{}}
			\label{table-abla-buildingseg}
		\end{table}

For cross-domain building segmentation, we perform the bidirectional experiments on the proposed cross-domain building segmentation dataset to verify the model performance, i.e., MBD$\rightarrow$IAILD and IAILD$\rightarrow$MBD.  In Table~\ref{table-abla-buildingseg}, taking MBD$\rightarrow$IAILD as an example, compared with \emph{Source only}'s 76.1 mIoU, our best model achieves 5.6\% improvement to 81.7\% mIoU. Here, it is worthy to mention that using ISIA and AIM separately can bring limited improvement compared with the model using IMA+GFA. We think it is because most of the building instances are densely arranged and vary hugely in appearance, which makes the instance extraction hard. However, by using ISIA and AIM module simultaneously, the model achieves greater performance improvement. While for the IAILD$\rightarrow$MBD, our best model achieve 71.9\% mIoU, which is 10.1\% mIoU higher than the \emph{Source only} setting. As shown in Fig.~\ref{fig-visualization_building}, we run several DA models on the target domain and output visualization results. The \emph{Source only} model is confused on the target domain due to the large domain gap between source and target domains. Although IMA+GFA reduces domain gap from image-level and feature-level, the model's performance is still poor on account of the large amount pixel misclassification. As a comparison, the proposed model achieves a better domain adaptation effect on the target domain and effectively reduces pixel misclassification. 

\begin{table}[h]
	\centering
	\small
	\caption{Ablation study on cross-domain road segmentation task.  IMA donates the image-level adaptation; GFA stands for global feature-level adaptation; ISIA is the proposed inter-class separation and intra-class aggregation mechanism. AIM indicates the proposed adaptive-weighted instance matching strategy.}
	\setlength{\tabcolsep}{0.45mm}{
		{\begin{tabular*}{0.48\textwidth}{@{\extracolsep{\fill}}c|l|cccc|ccc@{}}
				\toprule[0.4mm]
				%\midrule
				Task&Method&IMA&GFA&ISIA&AIM&Build.&Bg.&mIoU(\%) \\
				\midrule
				\multirow{7}{*}{\makecell{MRD\\$\downarrow$\\DeepGlobe}}&\emph{Source only}&&&&&24.8&96.1&60.4\\
				\cline{2-9}
				&+IMA~\cite{BidirectionalLF}&\checkmark&&&&28.5&95.8&62.1\\
				&+GFA~\cite{AdaptSegNet}&\checkmark&\checkmark&&&30.7&96.1&63.4\\
				&+ISIA&\checkmark&\checkmark&\checkmark&&\textbf{34.2}&\textbf{96.1}&\textbf{65.2}\\ 
				&+AIM&\checkmark&\checkmark&&\checkmark&30.4&95.8&63.1\\
				&+\emph{all}&\checkmark&\checkmark&\checkmark&\checkmark&31.8&95.9&63.9 \\
				\cline{2-9}
				&\emph{Target only}&&&&&38.9&97.8&68.3\\
				\midrule
				\multirow{7}{*}{\makecell{DeepGlobe\\$\downarrow$\\MRD}}&\emph{Source only}&&&&&30.6&93.8&62.2\\
				\cline{2-9}
				&+IMA~\cite{BidirectionalLF}&\checkmark&&&&33.0&95.2&64.1\\
				&+GFA~\cite{AdaptSegNet}&\checkmark&\checkmark&&&34.0&94.8&64.4\\
				&+ISIA&\checkmark&\checkmark&\checkmark&&\textbf{37.1}&\textbf{95.2}&\textbf{66.1}\\
				&+AIM&\checkmark&\checkmark&&\checkmark&36.2&94.6&65.4\\
				&+\emph{all}&\checkmark&\checkmark&\checkmark&\checkmark&36.7&95.1&65.9\\
				\cline{2-9}
				&\emph{Target only}&&&&&42.9&97.5&70.2 \\
				\bottomrule[0.4mm]
		\end{tabular*}}{}}
	\label{table-abla-roadseg}
\end{table}

\subsubsection{Cross-domain road segmentation}
For cross-domain road segmentation, we also perform the bidirectional experiments on the proposed cross-domain road segmentation dataset to verify the model performance, i.e., MRD$\rightarrow$DeepGlobe and DeepGlobe$\rightarrow$MRD. In Table~\ref{table-abla-roadseg}, taking MRD$\rightarrow$DeepGlobe as an example, the gap between the \emph{Source only} model the \emph{Target only} model is 7.9\% mIoU. By using the IA and GFA strategies, the adapted model achieves 3.0\% mIoU improvement  to 63.4\% on the target domain. By using the proposed ISIA, the adapted model achieves 65.2\% mIoU. However, when applying the AIM strategy on this task, the model's performance on the target domain dropped evidently. We think the reason lays on that it is hard to extract a instance for road targets, which is because they are usually connected to each other. On the other hand, since the large slenderness ratio of road targets, the down-sampling operation in the feature extraction network will cause the loss of target semantic features, which will result in poor segmentation performance. This phenomenon can be also seen in DeepGlobe$\rightarrow$MRD task.

\subsubsection{Comparison}
We conduct comparative experiments with SOTA models~\cite{AdaptSegNet, CLAN, UDADT}  on cross-domain remote sensing datasets. As seen in Table~\ref{table-comparative-Building} and Table~\ref{table-comparative-Road},  our method achieves the highest mIoU on MBD$\rightarrow$IAILD and MRD$\rightarrow$DeepGlobe. 
\begin{table}[h]
	\begin{minipage}{0.45\linewidth}
		\centering
		\caption{Segmentation accuracy comparison on cross-domain building segmentation.}
		\label{table-comparative-Building}
		\setlength{\tabcolsep}{1pt}{
			{\begin{tabular*}{1.0\textwidth}{@{\extracolsep{\fill}}c|c|c@{}}
					\toprule[0.4mm]
					Task   & Method   & mIoU(\%)  \\
					\midrule
					\multirow{4}{*}{\makecell{MBD\\$\downarrow$\\IAILD}} & Adaptseg~\cite{AdaptSegNet}  &78.5  \\
					\cline{2-3}
					&CLAN~\cite{CLAN} &79.1 \\
					\cline{2-3}
					&Wang et al.~\cite{UDADT} &81.1\\
					\cline{2-3}
					&Ours &\textbf{81.7} \\
					\bottomrule[0.4mm]
			\end{tabular*}}{}}
	\end{minipage}
	\quad
	\begin{minipage}{0.45\linewidth}  
		\centering
		\caption{Segmentation accuracy comparison on cross-domain road segmentation.}
		\label{table-comparative-Road}
		\setlength{\tabcolsep}{1pt}{
			{\begin{tabular*}{1.1\textwidth}{@{\extracolsep{\fill}}c|c|c@{}}
					\toprule[0.4mm]
					Task   & Method   & mIoU(\%)  \\
					\midrule
					\multirow{4}{*}{\makecell{MRD\\$\downarrow$\\DeepGlobe}} & Adaptseg~\cite{AdaptSegNet}  &61.9  \\
					\cline{2-3}
					&CLAN~\cite{CLAN} &62.6\\
					\cline{2-3}
					&Wang et al.~\cite{UDADT} &63.1 \\
					\cline{2-3}
					&Ours &\textbf{66.1 }\\
					\bottomrule[0.4mm]
			\end{tabular*}}{}}
	\end{minipage}
\end{table}

\subsection{Ablation Study}
\subsubsection{Discussion on impact of segmentation model}
While most current DA methods for domain adaptive segmentation use DeepLab-v2~\cite{DeepLabv2} as the segmentation model. However, how the capability of segmentation model affects the domain adaptation has not been explored. By using various universally effective segmentation models, we aim to reveal the relationship between domain adaptation strategy and segmentation model performance. As seen in Table.~\ref{table-abla-segnet}, three widely-used semantic segmentation models~\cite{DeepLabv2, DeepLabv3+, HRNet} are used for the domain adaptive segmentation task. For revealing the effectiveness of domain adaptation strategies, we propose a new metric called \emph{Normalized Adaptability Measure (NAM)} as follows:
\begin{equation}
	NAM = \frac{IoU_{Ada}-IoU_{SO}}{IoU_{TO}-IoU_{SO}} \times100\%
\end{equation}
where \emph{NAM} indicates the improvement of the adapted model performance against the source only setting. Intuitively, a large \emph{NAM} metric manifests a better adapation efficiency. \emph{TO} , \emph{SO} and \emph{Ada} represent target only setting, source only setting and the adapted model, respectively. 

As shown in Table~\ref{table-abla-segnet}, we conduct three cross-domain segmentation tasks including pixel-level annotation on street scenes, remote sensing building segmentation and road segmentation, respectively.  By analyzing the \emph{NAM} metric of each segmentation model, we found that as the performance of the segmentation model improves, the improvement brought by the domain adaptation strategy will gradually increase. This shows that when the learning ability of a segmentation model is strong enough, it can also cover the domain variant to a certain extent without any other adaptation strategy. Here we propose an assumption that for the case where the difference between domains is small, the segmentation model with good performance is enough to cover most of the domain gap; for the case of large differences between domains, the segmentation model with better performance tends to overfit in the source domain, and underfit in the target domain. This is because the model will pay more attention to the different features belonging to the source domain but not the target domain. For example, we take the GTA5$\rightarrow$Cityscapes as a \emph{hard} domain adaptation task, because there are multiple semantic categories and large intra-class differences. As seen in Table~\ref{table-abla-segnet}, experimental results demonstrate that when the performance of the segmentation model is enhanced, the performance improvement brought by the domain adaptation strategy is relatively strengthened since the \emph{NAM} metric increases. While for cross-domain building segmentation task, i.e., MBD$\rightarrow$IAILD, although the image resolutions of the source and target domains are different, the object appearance variance is small and there are only two categories, which can be taken as a \emph{simple} domain adaptation task. Thus even the \emph{target only} performance of FCN with HRNet-w48~\cite{HRNet} is better than that of DeepLab-v3+~\cite{DeepLabv3+}, the adapted performance of \emph{NAM} metric in the target domain is inferior than that of DeepLab-v3+. This phenomenon is also being observed in cross-domain road segmentation task, i.e., MRD$\rightarrow$DeepGlobe. Here, it is worthy mentioning that \emph{NAM} metric only evaluates the relative improvement of adapted model against the non-adapted model. Because the absolute performance of the adapted model increases with the performance of the segmentation model.
\begin{figure}[t]
	\centering
	\includegraphics[scale=0.32]{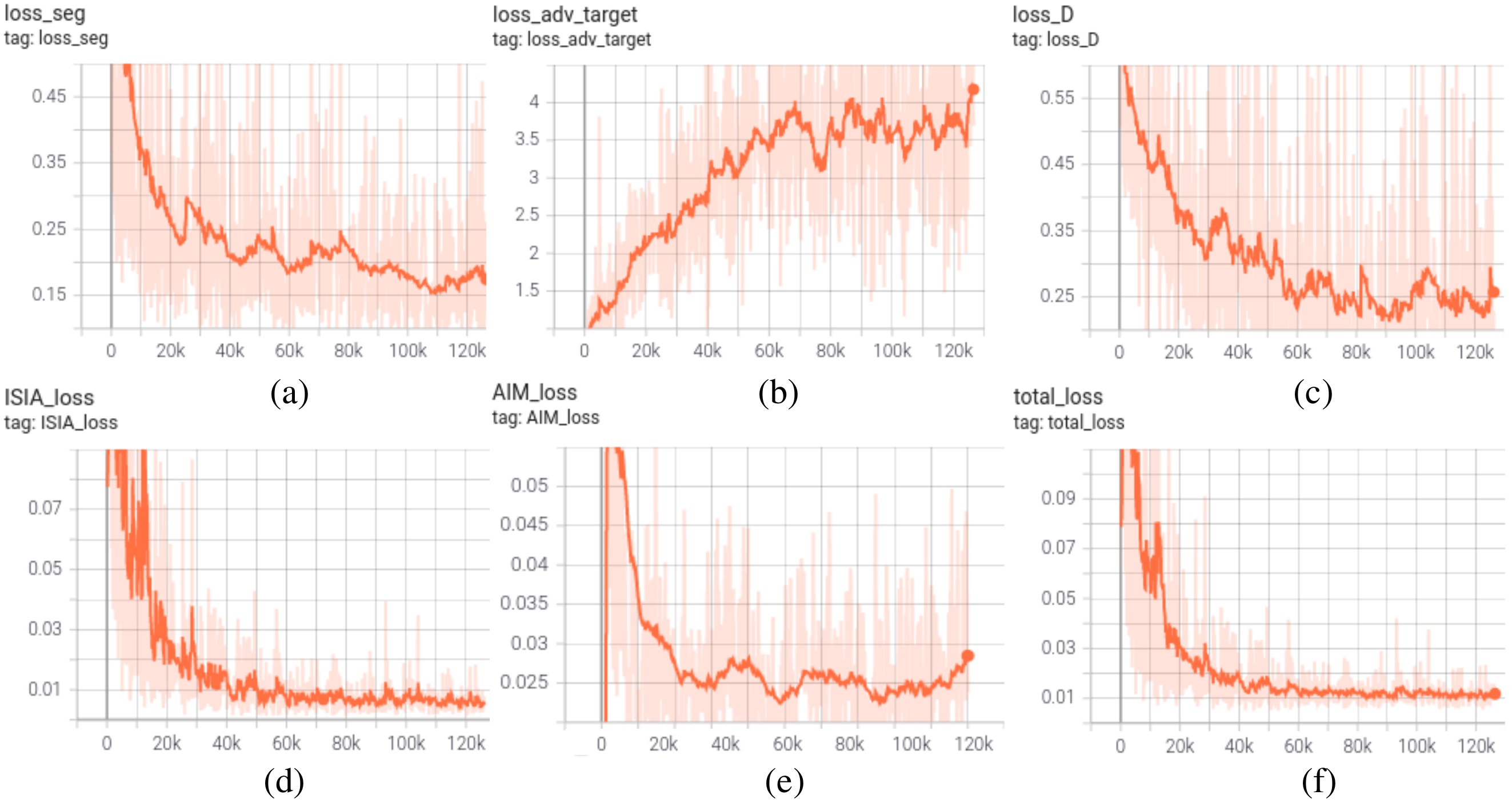}
	\caption{Loss visualizations of the proposed model on GTA5$\rightarrow$Cityscapes. (a) segmentation loss; (b) adversarial loss; (c) discriminant loss; (d) ISIA loss; (e) AIM loss; (f) the total loss.}
	\label{fig-loss_visualizations}
\end{figure}
\begin{table}[h]
	\centering
	\small
	\caption{Impact of segmentation model performance on domain adaptation task in mean IoU rate (\%). }
	
	\setlength{\tabcolsep}{0.45mm}{
	{\begin{tabular*}{0.48\textwidth}{@{\extracolsep{\fill}}c|c|c|ccc|c@{}}
	\toprule[0.4mm]
	Task &Seg.-Model &Backbone &\makecell[c]{Sour.\\only} &Ours &\makecell[c]{Tar.\\only} &\makecell[c]{NAM}\\
	\midrule
	\multirow{3}{*}{\makecell[c]{GTA5\\$\downarrow$\\Cityscapes}}&DeepLabv2~\cite{DeepLabv2}&ResNet-101&36.6&50.7&65.1 &48.1\\
	\cline{2-7}
	&DeepLabv3+~\cite{DeepLabv3+}&ResNet-101&46.8&66.3&78.4 &61.7 \\
	\cline{2-7} 
	&FCN~\cite{HRNet}&HRNet-w48&60.3&73.8&80.9 &\textbf{63.7} \\
	\midrule
	\multirow{3}{*}{\makecell[c]{MBD\\$\downarrow$\\IAILD}}&DeepLabv2~\cite{DeepLabv2}&ResNet-101&64.2&69.1&73.9&50.5\\
	\cline{2-7}
	&DeepLabv3+~\cite{DeepLabv3+}&ResNet-101&76.1&81.7&82.6&\textbf{85.9}\\
	\cline{2-7} 
	&FCN~\cite{HRNet}&HRNetv2-w48&78.4&83.0&84.3&78.0\\
	\midrule
	\multirow{3}{*}{\makecell[c]{MRD\\$\downarrow$\\DeepGlobe}}&DeepLabv2~\cite{DeepLabv2}&ResNet-101&49.8&53.8&58.5&46.6\\
	\cline{2-7}
	&DeepLabv3+~\cite{DeepLabv3+}&ResNet-101&60.4&65.2&68.3&\textbf{60.2}\\
	\cline{2-7} 
	&FCN~\cite{HRNet}&HRNet-w48&65.7&71.2&75.0&59.1 \\
	\bottomrule[0.4mm]
	\end{tabular*}}{}}
	\label{table-abla-segnet}
\end{table}

\subsubsection{Training stability}

Since the proposed model proceeds domain adaptation on multiple levels, i.e., image-level, feature-level, category-level and instance-level, as the loss functions consists of four components, which are segmentation loss, adversarial loss, ISIA loss and AIM loss.  We explore the stability of the training process. As shown in Fig.~\ref{fig-loss_visualizations}, the segmentation loss tends to converge with iterations increasing, which indicates that the model are adapted to both the source and target domains. While the generator loss rises and discriminator loss decreases that reveals the model's feature extraction ability increases. And we see that the ISIA loss is steadily decreasing that proves the proposed ISIA strategy towards continuous optimization. While for AIM loss, it has a warm up strategy for accurate instance extracting and then the loss decreases to a small-scale fluctuating state rapidly. We think it is because the accuracy of instance extraction is limited by the segmentation model, and the instance quantity varies in different images. And for the total loss, as it is a combination of multiple losses, it towards convergence which reveal the proposed model is able to adapted to the target domain. 

\subsubsection{Inter-class Separation vs. Intra-class Aggregation}
The proposed ISIA strategy performs inter-class separation and intra-class aggregation simultaneously. To reveal the contributions of both mechanisms, we design an ablation study as shown in Table~\ref{table-abla-ISvsIA}. The model with the proposed ISIA is observably better than that w/o. ISIA. Among all three domain adaptive segmentation tasks, the contribution of IA is a bit greater than IS, but both can bring significant performance improvement compared to~\cite{AdaptSegNet}. While the IS and IA can work together to achieve a better performance. As a result, we believe that pulling feature distributions of the same class across domains closer and pushing feature distributions of different classes across domains further are both beneficial to domain adaptation task. 
\begin{table}[h]
    \centering
    \small
    \caption{Contributions of IS vs. IA on cross domain segmentation task in mean IoU rate (\%).}
	\setlength{\tabcolsep}{0.45mm}{
	{\begin{tabular*}{0.48\textwidth}{@{\extracolsep{\fill}}c|clll@{}}
    \toprule[0.4mm]
    Task&IMA+GFA&+IS&+IA&+ISIA \\
    \midrule
    GTA5$\rightarrow$Cityscapes&45.3&47.4$_{\textcolor{red}{+2.1}}$&47.9$_{\textcolor{red}{+2.6}}$&49.6$_{\textcolor{red}{+4.3}}$\\
    \midrule
    MBD$\rightarrow$IAILD &80.9&81.1$_{\textcolor{red}{+0.2}}$&81.1$_{\textcolor{red}{+0.2}}$&81.3$_{\textcolor{red}{+0.4}}$ \\
    \midrule
    MRD$\rightarrow$DeepGlobe &63.4&64.3$_{\textcolor{red}{+0.9}}$&64.6$_{\textcolor{red}{+1.2}}$&65.2$_{\textcolor{red}{+1.8}}$ \\
    \bottomrule[0.4mm]
	\end{tabular*}}{}}
    \label{table-abla-ISvsIA}
\end{table}
 
\subsubsection{Computational Complexity}
Here shows the computational complexity of the proposed unsupervised semantic segmentation model. In essence, the proposed UDA method only change the distribution of parameters but does not change the FLOPs and complexity of the semantic segmentation model, we choose the representative domain adaptive segmentation task, i.e., GTA5$\rightarrow$Cityscapes, to calculate the FPS of our method. The experimental results are shown in Table~\ref{table-abla-complexity}.

\begin{table}[h]
	\centering
	\small
	\caption{Computational complexity.}
	\setlength{\tabcolsep}{0.45mm}{
		{\begin{tabular*}{0.48\textwidth}{@{\extracolsep{\fill}}c|cccc@{}}
				\toprule[0.4mm]
				Task&Param.&FLOPs&Memory&FPS \\
				\midrule
				GTA5$\rightarrow$Cityscapes&42.72M&183.92G&2441.78MB&12.81\\
				\bottomrule[0.4mm]
		\end{tabular*}}{}}
	\label{table-abla-complexity}
\end{table}

\subsubsection{Failure Analysis}
\begin{figure}[htbp]
	\centering
	\includegraphics[scale=0.45]{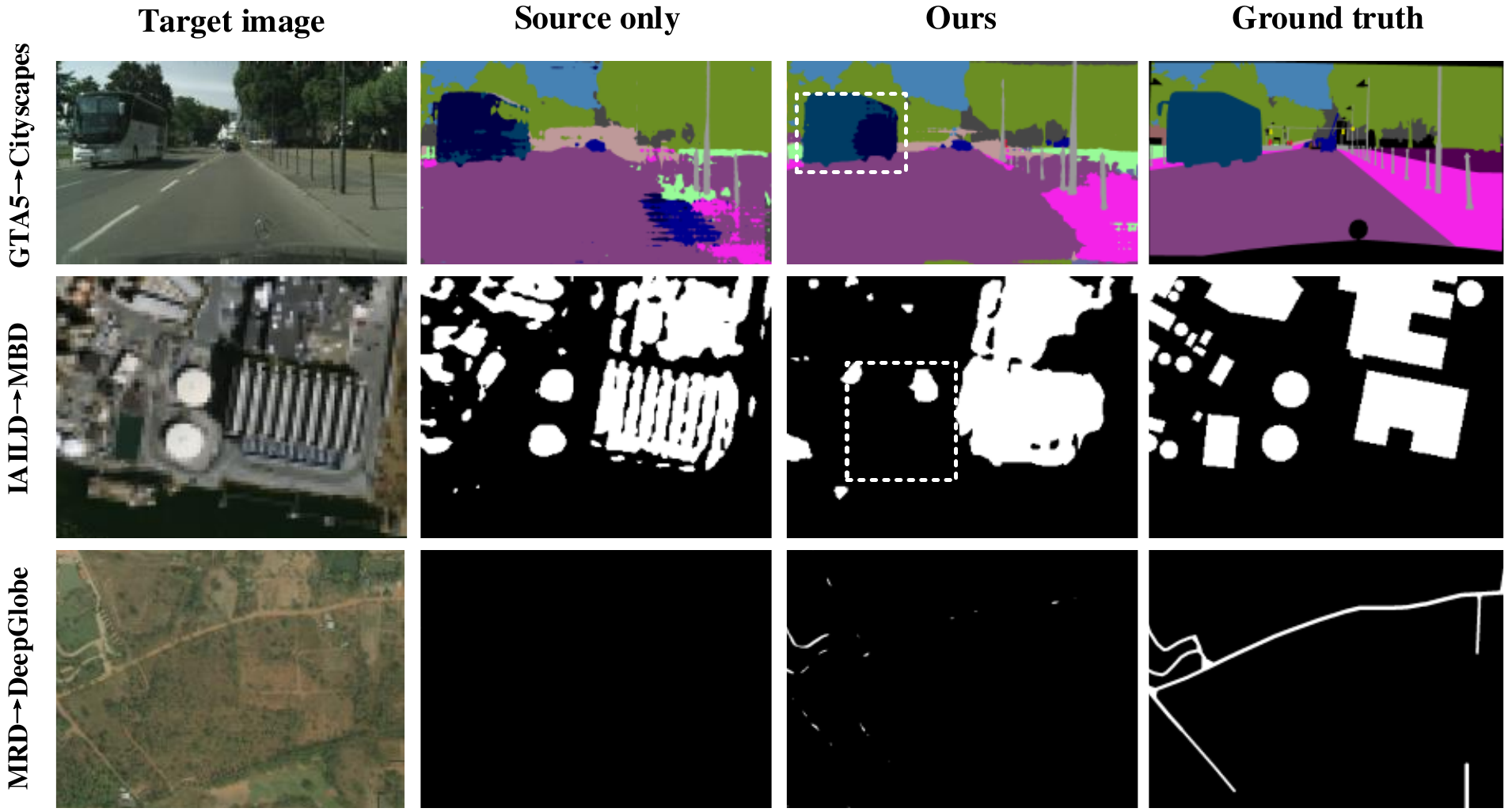}
	\caption{Failure cases from three cross-domain semantic segmentation tasks.}
	\label{fig-failure_case}		
\end{figure}

As seen in Fig.~\ref{fig-failure_case}, the proposed model may show less capability in such cases: 1) Complex inter-class similarity. For example, \emph{bus} and \emph{truck} have the similar visual appearance in GTA5$\rightarrow$Cityscapes. Even with the proposed UDA method, there is still prediction error between these two categories limited by the feature discrimination capability of the semantic segmentation model. Of course it is worth mentioning that there is an improvement in comparison to \emph{Source only} condition. 2) Large intra-class variation. For instance, in IAILD$\rightarrow$MBD task, all foreground objects are labeled as one category with great difference in shape, texture, gray scale, etc. The proposed model may tend to arise pixel misclassification. 3) Bottleneck of the semantic segmentation model. We believe it may be solved by using better semantic segmentation approaches.

%-----Conclusion Section-----
\section{Conclusion}
In this paper, we propose a multi-level unsupervised domain adaptation framework for cross-domain semantic segmentation which considers category homogeneity and diversity in the meantime. Thus the model can alleviate the class confusion problem by driving intra-class features closer and inter-class features further apart. Based on the alignment complexity of each category, we design an effective instance-level alignment strategy to further enhance the adaptation validity on hard categories. Finally, the model is trained in a self-supervised way by generating the pseudo labels for the target domain. In addition, we carry out cross-domain semantic segmentation on remote sensing images to extend the domain adaptation application. This paper also explores the impact of segmentation model performance on domain adaptation efficiency. The experimental results prove the proposed method can effectively reduce pixels misclassification among confusable categories and achieve a new state-of-the-art segmentation accuracy on benchmark datasets. In the future work, the following work will be scheduled. On the one hand, the proposed UDA method can be embedded more semantic segmentation models. On the other hand, more cross domain semantic segmentation tasks are being explored. And we also attempt to extend the UDA to more complex open-world problems.

\ifCLASSOPTIONcaptionsoff
  \newpage
\fi

{\small
	\bibliographystyle{IEEEtran}
	\bibliography{egbib}
}

\end{document}